\documentclass[conference]{IEEEtran}
\usepackage{times}

\usepackage{caption}
\usepackage[numbers]{natbib}
\usepackage{multicol}
\usepackage[bookmarks=true]{hyperref}
\usepackage{amsmath, amssymb}
\usepackage{booktabs}
\usepackage{graphicx}
\usepackage{xcolor}
\usepackage{float}
\usepackage{mathrsfs} %
\usepackage{caption}
\usepackage{tabularray} 
\usepackage{tabularx} 
\usepackage{ragged2e} 
\usepackage{cuted}
\usepackage{array}      %
\usepackage{multirow}   %
\usepackage[utf8]{inputenc}
\usepackage{svg} 
\usepackage{algpseudocode}  
\usepackage{pifont}
\usepackage{amsfonts}
\usepackage{dsfont}
\usepackage{balance}
\usepackage{stfloats}
\hypersetup{hidelinks}

\newcommand{\R}{\mathbb{R}}

\newcommand{\I}{\mathcal{I}}

\begin{document}

\title{Discrete-Time Hybrid Automata Learning: Legged Locomotion Meets Skateboarding}



\author{
    Hang Liu\textsuperscript{1} \quad
    Sangli Teng\textsuperscript{1}† \quad
    Ben Liu\textsuperscript{2} \quad
    Wei Zhang\textsuperscript{2} \quad
    Maani Ghaffari\textsuperscript{1} \\
    \textsuperscript{1}University of Michigan \quad
    \textsuperscript{2}Southern University of Science and Technology \\
    †Corresponding Author: {sanglit@umich.edu} \\
    Website, Code: 
    \href{https://umich-curly.github.io/DHAL/}{\texttt{\textcolor{magenta}{https://umich-curly.github.io/DHAL/}}}
}



%

\makeatletter
\let\@oldmaketitle\@maketitle
    \renewcommand{\@maketitle}{\@oldmaketitle
    \centering
    \includegraphics[width=1.0\textwidth]{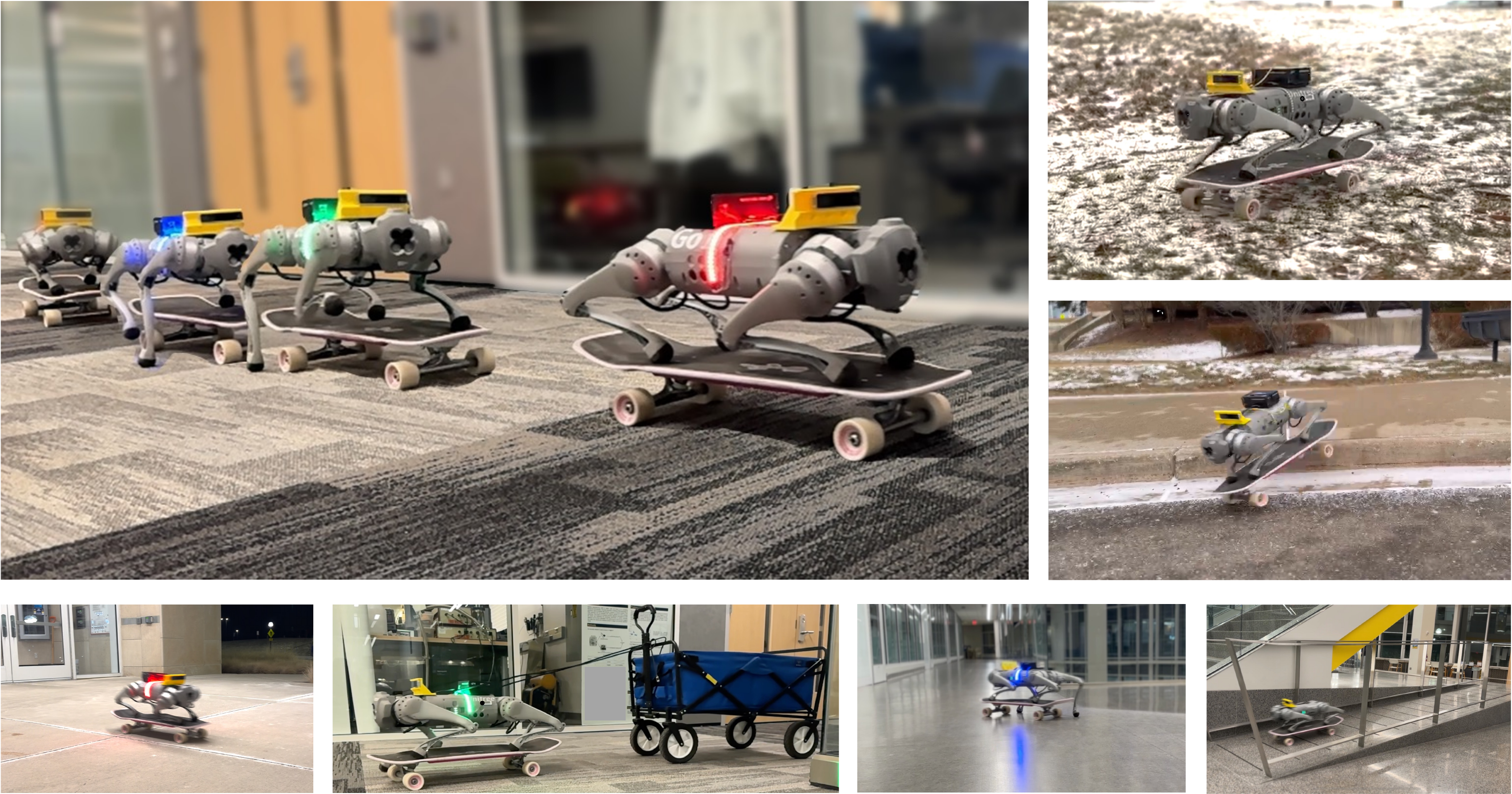}
    \vspace{-0.5cm}
    \captionof{figure}{Demonstration of DHAL performance across various indoor and outdoor terrains, including slopes, carpets, sidewalks, step, and scenarios with additional payloads or disturbance. The controller enables the robot to perform smooth and natural skateboarding motions, with reliable mode identification and transitions under disturbances.}
    \vspace{-0.2cm} 
    \label{fig:cover}
    \setcounter{figure}{1}
  }
\makeatother

\maketitle


\IEEEpeerreviewmaketitle



           
        
        

\begin{abstract} 
 Hybrid dynamical systems, which include continuous flow and discrete mode switching, can model robotics tasks like legged robot locomotion. Model-based methods usually depend on predefined gaits, while model-free approaches lack explicit mode-switching knowledge. Current methods identify discrete modes via segmentation before regressing continuous flow, but learning high-dimensional complex rigid body dynamics without trajectory labels or segmentation is a challenging open problem. This paper introduces Discrete-time Hybrid Automata Learning (DHAL), a framework to identify and execute mode-switching without trajectory segmentation or event function learning. Besides, we embedded it in reinforcement learning pipeline and incorporates a beta policy distribution and a multi-critic architecture to model contact-guided motions, exemplified by a challenging quadrupedal robot skateboard task. We validate our method through sufficient real-world tests, demonstrating robust performance and  mode identification consistent with human intuition in hybrid dynamical systems.

\end{abstract}

\section{INTRODUCTION}

In state space, systems that exhibit both flow-based continuous dynamics and jump-based discrete dynamics are known as hybrid dynamical systems\cite{branicky2005introduction}. Such systems are prevalent in many real-world environments, particularly those involving discontinuities, contact events, or mode switching---examples include legged robotics, power systems (e.g., thermostat systems\cite{alur2003counter}, DC-DC converters), and even neurobiological models such as the integrate-and-fire neuron\cite{touboul2009spiking}.

In the field of robotics, the walking behavior of bipedal robots is a quintessential example of a hybrid dynamical system. In model-based control, the Hybrid Automaton framework has been proposed as a powerful tool to describe systems that encompass both discrete and continuous dynamics~\cite{1166523, sreenath2011compliant}. This framework has been widely adopted for behavior planning~\cite{khazaee2021behavior} and legged locomotion. However, these approaches typically rely on simplified dynamic models, and contact events are often pre-defined, as in manually crafted walking gaits.

With the rise of data-driven paradigms and world models, recent research has begun exploring learning-based approaches for hybrid dynamics\cite{JMLR:v13:ly12a, chenlearning, NEURIPS2021_5291822d}. However, most of these efforts remain in the exploratory stage, typically focusing on low-dimensional, simple systems or single trajectory scenarios, and non-efficient training. Developing a method that can generalize to high-dimensional systems, handle complex real-world dynamics, and effectively identify discrete modes remains an open problem.

To address this, we propose a generalized approach for mode identification and prediction in hybrid dynamical systems. We design a set of hybrid dynamics modules that integrate discrete hybrid automata with a variational autoencoder (VAE) framework, enabling the system to heuristically learn both mode identification and flow dynamics. We evaluate our approach in contact-guided scenarios, which involve complex sequences of contact events~\cite{zhang2024wococo}, to demonstrate its effectiveness.

Designing such contact-rich scenarios is highly non-trivial. Unlike model-based methods, model-free reinforcement learning (RL) has shown promise in solving optimal control problems (OCPs) by modeling dynamics as a Markov Decision Process. Model-free RL requires minimal assumptions and can be applied to a wide range of tasks across diverse dynamical systems~\cite{li2021reinforcement, ji2022hierarchical}. To address contact-rich tasks, we integrate a multi-critic architecture with a Beta distribution policy, and embed our hybrid dynamics system into the RL pipeline.

we tackle the challenging task of enabling a quadrupedal robot to skateboard. This task exemplifies a highly dynamic, contact-guided, and hybrid underactuated system, requiring precise handling of mode transitions and contact events. The key contributions of this work are summarized as follows:

\begin{enumerate}
    \item \textbf{Discrete Hybrid Automata Framework}: We propose a discrete hybrid automata framework for online reinforcement learning that eliminates the need for explicit trajectory segmentation or event labeling.
    \item \textbf{Contact-guided task design}: We combined multi-critic architecture and beta distribution to effectively address contact-guided problem in hybrid systems.
    \item \textbf{Sim2Real of Underactuated Skateboarding motion}: We achieved agile and robust sim-to-real performance in the highly underactuated and hybrid task of skateboard motion.
\end{enumerate}

\begin{figure*}[t]
    \centering
    \includegraphics[width=1\linewidth]{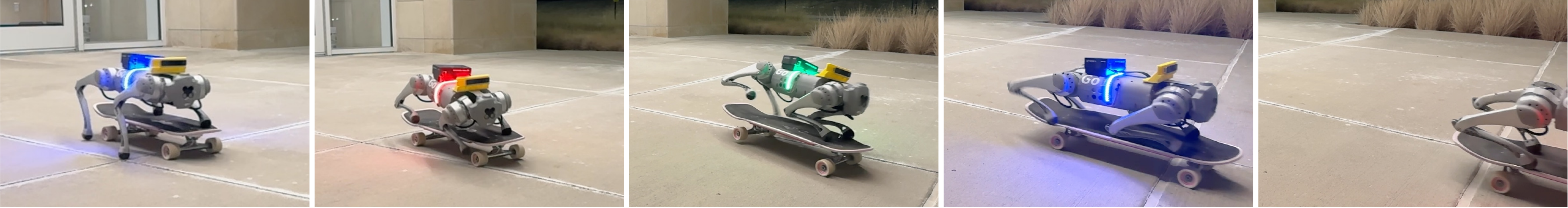}
    \caption{The potholes on the downhill slope caused the robot dog's right front leg to get stuck, preventing it from smoothly getting onto the board. In fact, both of its hind legs even lost contact with the board. Nevertheless, the policy was still able to guide the robot dog to jump back onto the board and complete the recovery behavior.}
    \label{fig:RobotRecover}
\end{figure*}

\section{RELATED WORK}

\subsection{Legged Robot Control}
The Model Predictive Control (MPC)~\cite{teng2022lie, teng2024convex,Teng-RSS-23} with simplified single rigid body has been successfully applied to motion planning of legged robot\cite{corbères2024perceptivelocomotionwholebodympc, 10138309, teng2022error, teng2021toward}, achieving robust locomotion on flat ground under diverse gait patterns. \citet{corbères2024perceptivelocomotionwholebodympc, 10138309} further integrated motion planning and perception, enabling quadruped robots to navigate complex terrains. However, such approaches are highly dependent on accurate global state estimation, which poses limitations in outdoor and long-range scenarios. Additionally, the gait pattern of model-based approach are designed manually, which can not scale in complicated scenarios.

In contrast to the model-based approach, model-free reinforcement learning (RL) has demonstrated remarkable capabilities in legged robot control, including high-speed locomotion \cite{margolis2024rapid, 10609983, he2024agile, li2024reinforcement}, complex terrain traversal \cite{cheng2024extreme, cheng2024quadruped, gu2024advancing, 10161144}, manipulation, and interaction \cite{fu2023deep, liu2024visual, ji2023dribblebot}. The main focus of RL-based quadruped control in recent years is in the paradigm of sim-to-real transfer \cite{kumar2021rma,gu2024advancing}, safe reinforcement learning \cite{he2024agile}, and gait control \cite{margolis2023walk, yang2022fast, kim2024learning}. In this paper, we primarily focus on the exploration of sparse motion patterns and the embedding of hybrid dynamics learning.

\subsection{Contact-guided locomotion pattern}
The hybrid nature of contact dynamics makes it challenging to synthesize optimal motions for contact-rich tasks. In model-based methods, the discontinuities introduced by contact create obstacles for gradient-based optimization. \citet{doi:10.1177/02783649241273645} and \cite{jin2024task} address this issue by formulating it as a linear complementarity problem (LCP) and relaxing the complementarity constraints, while \citet{1166523} adopt a hybrid dynamic system combined with automata to handle contact. 

In the realm of reinforcement learning (RL), the specification of sparse or contact-guided motion patterns has not been widely investigated or solved, such as explicitly prescribing contact-based motion \cite{zhang2024wococo} or relying solely on key frame trajectory tracking \cite{zargarbashi2024robotkeyframing}. Considering real-world robot collisions and constraints, sparse reward design in a high-dimensional sampling space severely limits effective exploration \cite{zhang2024wococo}. In contrast, under a multi-critic framework \cite{zargarbashi2024robotkeyframing} combined with a Beta action distribution \cite{chou2017improving}, this paper achieves contact-specified “skateboarding” motion on a quadruped robot, with on-the-fly adjustments of the gait pattern.

\subsection{Hybrid Dynamics}

Hybrid dynamic systems are employed to describe systems that feature both continuous states and discrete modes, and they are widely used in model-based control and cyber-physical systems \cite{xiao2021robotic}. For instance, floating-base robots are often treated as hybrid dynamic systems due to the discontinuities and jumps caused by contact. Recently, \citet{JMLR:v13:ly12a, chenlearning, NEURIPS2021_5291822d, teng2024generalized} have leveraged data-driven approaches to construct either discrete or continuous hybrid dynamic systems.

Unlike previous work, we integrate reinforcement learning with hybrid dynamics, enabling the robot to learn a Hybrid Dynamics Automaton for explicit mode switching and control without requiring labels or segmentation.
\begin{figure*}[t]
    \centering
    \includegraphics[width=1\linewidth]{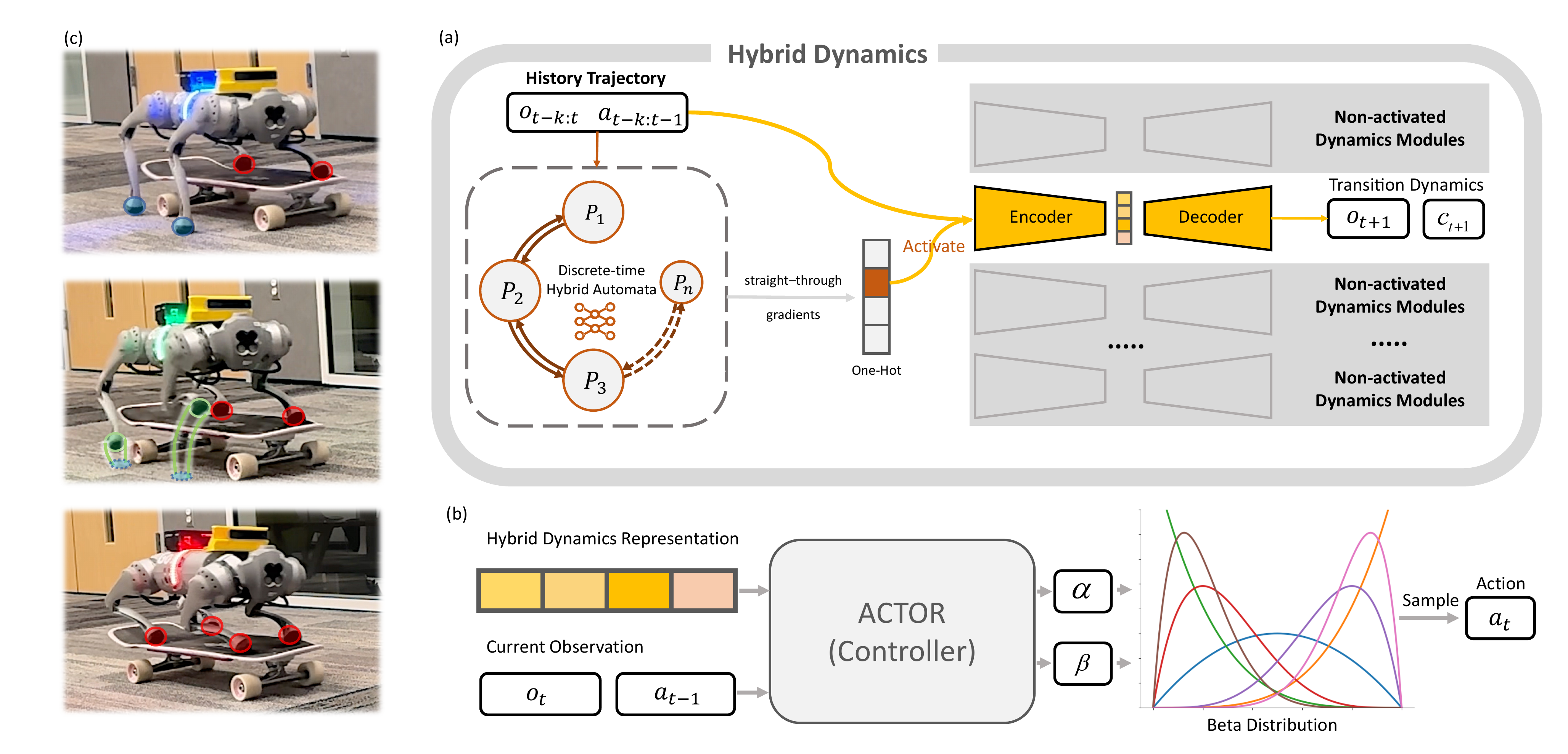}
    \caption{Discrete-time Hybrid Dynamics Learning (DHAL) Framework: (a) During training, the network learns to select the mode and activate the corresponding dynamics module (yellow-highlighted) to predict transition dynamics and contact. Here, $P_i$ represents the probability of the robot being in mode $i$ at time $t$. (b) The temporal features extracted by the encoder are combined with the current state and last action into the actor. The actor update $\alpha,\beta$, which define the probability density function of the Beta distribution, and then samples joint actions from the Beta distribution. (c) In a real-world deployment, we use different LED colors to indicate the active modes, showcasing smooth transitions and mode-specific behaviors.}
    \label{fig:framework}
\end{figure*}

\section{BACKGROUND AND PROBLEM SETTING}
For clarity and reference, Table \ref{tab:symbol} provides an overview of the key symbols and abbreviations used throughout this paper.

\subsection{Markov Decision Process}
We model the robot control problem as an infinite-horizon partially observable Markov decision process (POMDP), composed by tuple $\mathcal{M} = (S, \mathcal{O}, A, p, r, \gamma)$, with $s_t \in \mathcal{S}$ the full states, $o_t \in \mathcal{O}$ the partial observation of the agents from the environment,  $a_t \in \mathcal{A}$ the action the agent can take to interact with the environments, and $p(s_{t+1}|s_t,a_t)$ the transition function of state $s_t$. Each transition is rewarded by a reward function $r: S \times A \to \mathbb{R}$ with $\gamma$ representing a discount factor. The Reinforcement Learning optimization objective is to maximize the expected total return $\mathbb{E} \left[ \sum_{t=0}^{T} \gamma^t r_t \right]$.

\subsection{Discrete Hybrid Dynamical System}
\label{sec:Background-B}
Though incorporating dynamic information into policies can potentially enhance performance\cite{10161144}, most existing approaches assume a single dynamics model for legged robots, overlooking their inherently hybrid nature. To address this limitation, we aim to integrate information of the hybrid dynamical systems into policy learning to achieve superior performance.

Hybrid dynamical system involves both discrete and continuous dynamics or states. The system has a continuous \emph{flow} in each discrete \emph{mode} and can jump between these modes \cite{goebel2009hybrid,NEURIPS2021_5291822d}. Although utilizing a set of ordinary differential equations (ODEs) with transition maps modeled by neural networks has shown some promising results \cite{NEURIPS2021_5291822d, chenlearning}, the continuous integration is computational consuming for RL and not well-compatible for real-world robotic control, which is a digital control system. In this work, inspired by \citet{borrelli2017predictive} and \citet{JMLR:v13:ly12a}, we adopt \emph{discrete hybrid automata} to model the dynamics of legged robots, which naturally exhibit hybrid behaviors.

For simplicity of embedding discrete hybrid automata into the reinforcement learning framework, we utilize the concept of a \emph{switched affine system} to model the hybrid dynamics for legged robots. However, we utilize a set of nonlinear functions instead of affine functions to model the dynamics in each mode for legged robots. The dynamics of the legged robot can be described as
\begin{equation}
    \begin{aligned}
        s_{t+1} &= f^{i_t}(s_t,a_t),\\
    \end{aligned}
    \label{eq:1}
\end{equation}
where $s_t\in\R^n$ is the states, $a_t\in \R^m$ is the input, $i_t\in\I=\{1,2,...,K\}$ is the modes at time step~$t$, $f^{i_t}:\R^n\times \R^m\to\R^n$ is the dynamics on mode $i$. The mode $i_t$ is determined by an extra \emph{mode selector}. We assume that all states of the system are continuous-valued, not considering any discrete-valued state. Dynamics for each mode is unique, i.e., $f^i\neq f^j,\forall i,j\in\I$. The maximum number of mode $K$ is assumed to be known. 

In this work, we use neural networks to model both the dynamics and the automata (mode selector), aiming to extract a latent representation that informs the actor. To maintain consistency with the stochastic nature of the MDP, we utilize a $\beta$-VAE to model the state transition in equation \eqref{eq:1}. Unlike hybrid systems described by continuous flow \cite{goebel2009hybrid,NEURIPS2021_5291822d}, we omit the explicit jump mapping between different modes, as it is captured within the discrete-time dynamics \eqref{eq:1}.

\subsection{Environment Design}
To demonstrate the effectiveness of our approach, we aimed to select a challenging environment that involves complex mode transitions and contact-rich dynamics. Inspired by the real-world example of dogs learning to ride skateboards, we identified the task of a robotic dog skateboarding as a highly demanding scenario. This task presents distinct hybrid dynamics challenges, such as the significant differences between the gliding mode and the pushing mode. We believe this represents a highly worthwhile and meaningful challenge for validating our method. Our method is not limited to the skateboarding task but can also be expanded to other scenarios.

In this paper, we conducted experiments using the Unitree Go1 robot. The skateboard used in our experiments measures 800 mm × 254 mm × 110 mm. The Unitree Go1 is equipped with 12 actuated joints. Although it lacks the spinal degrees of freedom present in real quadrupedal animals, our experimental results demonstrate that it can still effectively control the skateboard's movement. 

To simplify the design of the system, we connected the end of the Unitree Go1's left forelimb to the skateboard using a spherical joint in simulation, providing passive degrees of freedom along the x, y, and z axes. Unlike \citet{chen2019feedback}, The wheels of the skateboard can only passively rotate around their respective axes. Additionally, we simulated the skateboard's truck mechanism in the simulation using a position PD controller to replicate its mechanical behavior.

\begin{table}
    \centering
    \caption{Important symbols and abbreviations}
    \begin{tabular}{cc}
    \toprule
        \textbf{Meaning} & \textbf{Symbol}\\
        
        \midrule
        \multicolumn{2}{c}{\textbf{POMDP}} \\
        \midrule
        Full State & $s_t$ \\
        Partial Observation & $o_t$\\
        Action & $a_t$\\
        State Space & $\mathcal{S}$\\
        Action Space & $\mathcal{A}$\\
        Discount Factor & $\gamma$\\
        
        \midrule
        
        \multicolumn{2}{c}{\textbf{Hybrid Dynamics System}} \\
        \midrule
        Discrete-time dynamics & $f^{i_t}$\\
        Mode index & $i_t$\\
        Number of modes & $K$\\
        Jacobian & $J$ \\
        Probability of each mode & $p$\\
        mode indicator vector & $\delta $ \\
        maximum number of modes & $K = |\delta|$ \\
    
        \midrule
        \multicolumn{2}{c}{\textbf{Environment}} \\
        \midrule
        Joint Position & $q =  \left[q_{Hip,Thigh,Calf}^{FL/FR/RL/RR}\right]$\\
        Joint Velocity & $\dot{q} =  \left[\dot{q}_{Hip,Thigh,Calf}^{FL/FR/RL/RR}\right]$\\
        Go1 Base Roll, Pitch,Yaw & $\phi, \theta, \psi$\\
        Go1 Base angular velocity & $\omega_{x}, \omega_{y}, \omega_{z}$\\
        Gravity & $g_{x,y,z}$\\
        Action & $a =  \left[a_{Hip,Thigh,Calf}^{FL/FR/RL/RR}\right]$\\
        Phase & $\Phi$\\
        Command & $cmd=[c_x, c_{yaw}]$\\
        Proprioception & $o=[q,\dot{q},g_{x,y,z},\phi, \theta, \psi, \omega_{x}, \omega_{y}, \omega_{z}]$\\
        Contact & $c$ \\
        Torque & $\tau$\\

        \bottomrule
    \end{tabular}

    \label{tab:symbol}
\end{table}
\section{METHODS}

We introduce DHAL, as shown in Fig. \ref{fig:framework}, to illustrate the proposed controller ~\cite{1166523} that leverages the model-based hybrid dynamics. More specifically:
\begin{enumerate}
    \item[\textit{i.}] \textbf{Discrete-time Hybrid Automata}, a discrete mode selector, to identify at each step time the one-hot latent mode of the system.
    \item[\textit{ii.}] \textbf{Dynamics Encoder}, based on the mode \( z \) chosen from discrete-time hybrid automata, the chosen dynamics encoder will be activated and get a tight representation of flow dynamics.
    \item[\textit{iii.}] \textbf{Dynamics Decoder}, to decode the representation and predict transition dynamics \( o_{t+1} \) and contact event \( c_{t+1} \).
    \item[\textit{iv.}] \textbf{Controller}, based on the tight representation from the dynamics encoder and observation at the current moment, control robot to perform skateboarding.
\end{enumerate}

\subsection{Discrete Neural Hybrid Automata}
\begin{figure}[ht]
    \centering
    \includegraphics[width=0.75\linewidth,trim=250 150 250 150, clip,]{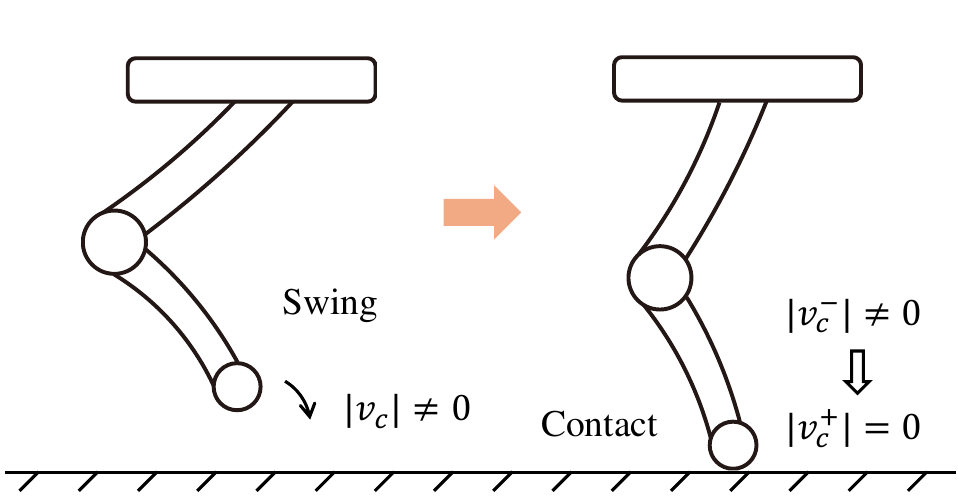}
    \caption{Switching of a hybrid system with inelastic collision. When the leg makes contact with the ground, the linear velocity will abruptly drop to zero.}
    \label{fig:contact-triggerl}
\end{figure}

Previous research has explored modifications to the framework of reinforcement learning, such as incorporating estimators to predict transition dynamics \cite{10161144}. However, prior work has not explicitly considered and addressed the issue of dynamics with mode switching. 
The dynamics of legged robots inherently exhibit hybrid behavior due to contact events, as illustrated in Fig. \ref{fig:contact-triggerl}. We model the contact as a perfect inelastic collision, which means the velocity of the contact point instantaneously drops to zero from a nonzero value. Consequently, the robot’s state also undergoes a discrete jump, resulting in a hybrid dynamics formulation.
To illustrate this, consider a simplified example of a single 3-DoF leg attached to a fixed base. The velocity of the contact point is given by
\begin{equation}
    v_c=J_c(q)\dot q,
\end{equation}
where $v_c\in\R^3$ is the linear velocity of the contact point, $J_c(q)\in\R^{3\times 3}$ is the Jacobian, $\dot q\in\R^3$ is the joint angular velocity. In most cases, $J_c(q)$ is invertible, implying a discontinuity in $v_c$ directly induces a discontinuity in $\dot q$. This simple example highlights the inherent hybrid nature of legged robot dynamics.

In this work, we aim to extract a latent representation of the dynamics for the controller. To achieve this, we model the dynamics for each mode and design a discrete-time hybrid automata (DHA) to determine which dynamic is active at any given time, as illustrated in Fig. \ref{fig:framework}. Since only partial observations $o_t$ are available in practice, rather than the full state $s_t$, we consider partial discrete-time dynamics. As discussed in Sec. \ref{sec:Background-B}, we employ a $\beta$-VAE instead of a deterministic dynamics model to better align with the stochastic nature of reinforcement learning. This process can be expressed by
\begin{equation}
    \begin{aligned}
        i_t &=f_\texttt{DHA}(o_{t-k:t}, a_{t-k-1:t-1};\theta_{\texttt{DHA}})\\
        o_{t+1},c_{t+1} &= f_\texttt{dec}^{i_t} \circ f_\texttt{enc}^{i_t}(o_{t-k:t}, a_{t-k-1:t-1};\theta_{\texttt{vae}}).
    \end{aligned}
    \label{eq:3}
\end{equation}
where $i_t\in\{1,2,...,K\}$ is the mode for current time $t$, which is determined by a state-action sequence \((o_{t-k:t}, a_{t-k-1:t-1})\). $c_{t+1}\in\mathcal S$ represents the contact information for each foot, see \eqref{eq:7}. The maximal number of modes is pre-defined. The next partial state $o_{t+1}$ and contact information $c_{t+1}$ is determined by a $\beta$-VAE at mode $i$, taking the same input as the hybrid automata. A key advantage of the $\beta$-VAE is that it directly provides a latent representation of the dynamics. In conventional switched affine systems \cite{10138309}, the hybrid automata relies on multiple modules. In contrast, our approach simplifies it to a single network that takes a window of past state-action as input.

Specifically, The hybrid automata is designed to output a one-hot latent, making this a classification problem. To determine the mode, we employ a combination of a softmax classifier and categorical sampling.
First, the probability distribution over modes is computed as
\begin{equation}
    p = f_{\texttt{softmax}}\circ f_{\texttt{DHA\_logit}}(o_{t-k:t}, a_{t-k-1:t-1};\theta_{\texttt{DHA}})
\end{equation}
where $p$ is the probability for each mode, $f_{\texttt{softmax}}$ is the softmax loss, $f_{\texttt{logit}}$ is the logit function, $\theta_{\texttt{DHA}}$ represents the parameters of the DHA neural network. Next, categorical sampling is applied to $p$ to obtain a one-hot latent vector $\delta$, which indicates the selected dynamics mode:
\begin{equation}
    \delta \sim \texttt{Categorical}(p), \quad \sum_{i=1}^{n} \delta_i = 1, \quad \delta_i \in \{0, 1\}
\end{equation}
where $\delta_i=1$ represents the mode is $i_t = i$.

With the hybrid automata, we build \( K \) corresponding dynamics $\beta$-VAEs for \( K \) modes. For each mode, the $\beta$-VAE encodes the historical trajectory and extracts the time-sequence feature into a latent representation, which will be subsequently decoded into the next partial state $o_{t+1}$. The latent representation in $\beta$-VAE encodes substantial dynamics information; hence, we then use it in the controller. For the encoder, we adopt a transformer architecture due to its superior ability to capture temporal dependencies, which can be expressed by
\begin{equation}
    z_t = \sum_{i=1}^{M} \delta_i \cdot f_{\texttt{enc}}^{i_t}(o_{t-k:t}, a_{t-k-1:t-1};\theta_{\texttt{enc}}),
\end{equation}
where $z_t\in\R^n$ is the latent representation, and the mode selector is included using $\delta$. The decoder then uses this representation to predict $o_{t+1}, c_{t+1}$, which can be expressed by
\begin{equation}
    \hat{o}_{t+1}, \hat{c}_{t+1} = \sum_{i=1}^{M} \delta_i \cdot  f_\texttt{dec}^{i_t}(z_t;\theta_{\texttt{dec}}),
    \label{eq:7}
\end{equation}
where $c_{t+1}\in[0,1]^n$ is the probability that each leg is in contact at time $t+1$. We include contact information to make learning hybrid switching easier for the networks. 

Typically, The ground truth for mode label and partial states are required to train these two neural networks \cite{NEURIPS2021_5291822d}; however, the mode label is hard to obtain even in simulation. To address this, motivated by \citet{10197536}, we utilize the unsupervised learning method to train the mode selector. We train the mode selector and the $\beta$-VAE simultaneously as \eqref{eq:3}, where the mode is self-determined by constructing the loss function as
\begin{equation}
\mathcal{L}_{\texttt{vae}} = \sum_{t}\texttt{MSE}(\hat{o}_{t+1}, {o}_{t+1}) + \texttt{BCE}(\hat{c}_{t+1}, {c}_{t+1}) + \beta \mathcal{L}_\mathcal{KL}), 
\end{equation}
where ${o}_{t+1},{c}_{t+1}$ are the ground truth values, $\texttt{MSE}$ represents mean-square-error loss, $\texttt{BCE}$ represents binary-cross-entropy loss. The KL divergence prevents the encoder from fitting the data too flexibly and encourages the distribution of the latent variable to be close to a unit Gaussian, while $\beta$ is used to control the trade-off scale. The hybrid automata is trained by minimizing the prediction error of $o_{t+1},c_{t+1}$, where the correct mode label will result in a lower prediction error. Such unsupervised learning does not require a ground truth for mode labels. The discrete categorical sample results in discontinuity of gradients, we apply the straight-through-gradient method \cite{bengio2013estimating} to realize backpropagation during the training. Moreover, the gradient of the discrete-time automata is independent of PPO, ensuring accurate mode identification. However, the gradient backpropagates through the encoder to extract useful temporal features.

In addition, we encourage the mode to be distinguished by minimizing the information entropy of the mode probability $p$, resulting in the final loss:
\begin{equation}
    \mathcal{L}_{\texttt{DHA}} = \mathcal{L}_{\texttt{vae}} + \mathcal{H}(p),
\end{equation}
where $\mathcal{H}$ represents information entropy, aiming to ensure that the modes' probabilities are as distinct as possible and to prevent confusion between different modes.

Compared to \cite{NEURIPS2021_5291822d} that models hybrid dynamics using a continuous flow approach, we adopt a discrete-time formulation. This formulation unifies mode switching and dynamics within a single framework rather than treating them separately. Furthermore, unlike \cite{NEURIPS2021_5291822d} that requires pre-segmentation of the trajectories to distinguish the mode, our method integrates mode selection directly into the training process, streamlining the entire workflow. The detailed network architecture and hyper-parameters can be found in the Appendix.

\subsection{Multi Critic Reinforcement Learning}

\begin{figure}
    \centering
    \includegraphics[width=1\linewidth]{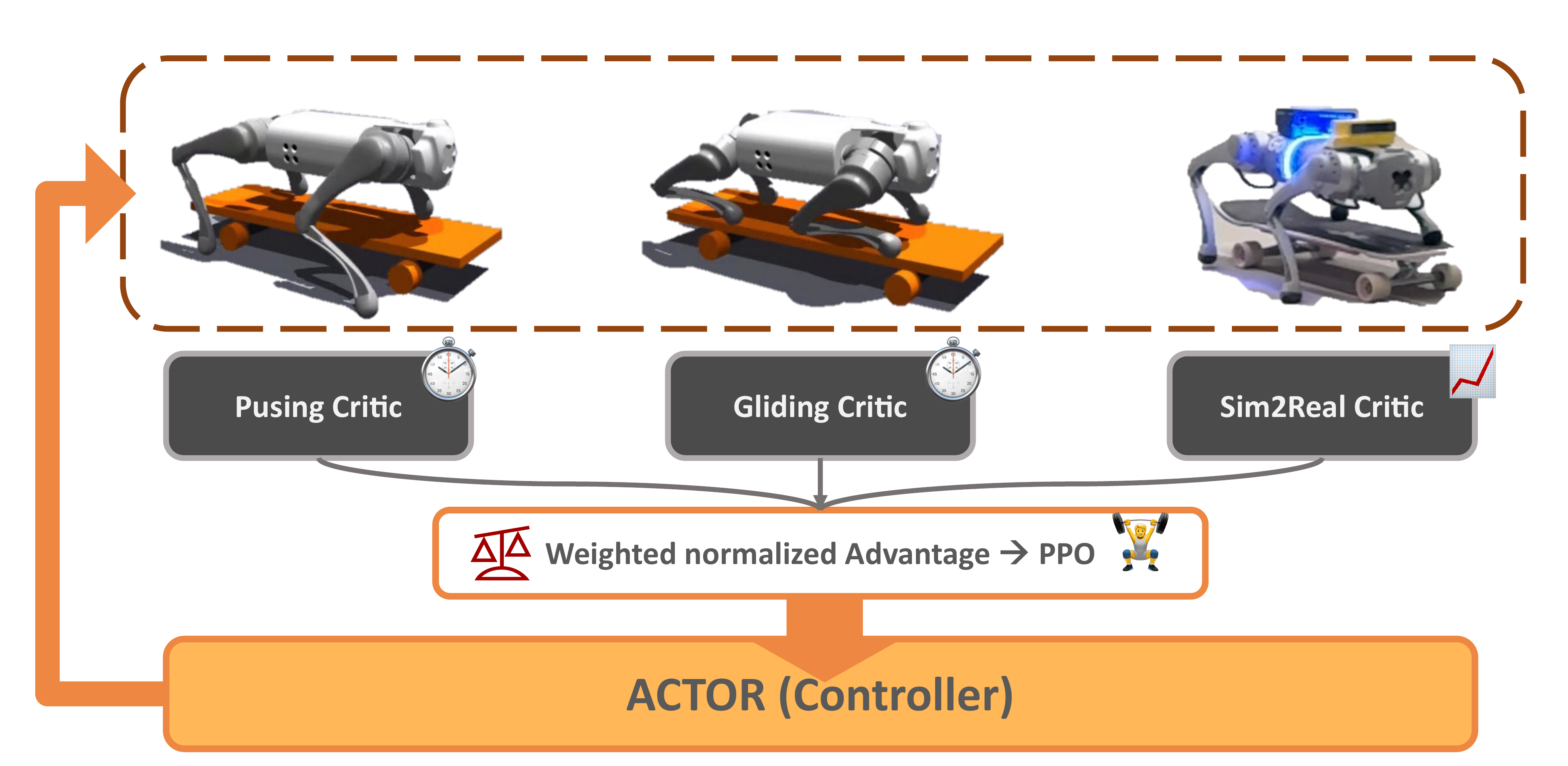}
    \caption{Multi-Critic Skateboard Task}
    \label{fig:multi-critic}
\end{figure}

Inspired by \citet{zargarbashi2024robotkeyframing,xing2024multi}, we adopt the concept of multi-critic learning and apply it to the design of multi-task objectives for different motion phases. Designing an intuitive reward function for the robotic dog to perform skateboarding maneuvers is challenging. Therefore, we leverage sparse contact-based rewards to guide the robot in executing smooth pushing motions and gliding on the skateboard—two distinctly different movement patterns. Unlike the approach in \citet{fuminimizing}, where minimizing energy consumption is prioritized, we found that this approach struggles to naturally induce the gliding motion. This difficulty arises because skateboarding, compared to standard quadrupedal locomotion, has a much smaller stability margin. As a result, the robot tends to avoid using the skateboard altogether to minimize energy loss rather than embracing it as part of the task.

To address this challenge, we design two distinct tasks corresponding to different phases of the cycle: \emph{gliding} and \emph{pushing}. During the pushing phase, the robot's primary objective is to track the desired speed, while during the gliding phase, its goal is to maintain balance and glide smoothly on the skateboard. Although we specify the motion phase transitions using cyclic signals during training to ensure balanced learning of both tasks, in real-world deployment, these cyclic signals can be adapted on the fly or generated based on the robot's velocity. The cyclic signal serves as an indicator to inform the robot with which movement pattern to adopt.

However, because the speed-tracking rewards and regularization terms for sim-to-real adaptation are dense, while the contact-related rewards are inherently sparse at the early stages of exploration, using a single critic to estimate the value of all rewards can lead to the sparse rewards' advantages being diluted by the dense rewards during normalization. This makes it difficult for the robot to learn the desired different style behaviors \cite{zargarbashi2024robotkeyframing}. 

To mitigate this and generate different style motions, we introduce a multi-critic framework, as illustrated in Fig. \ref{fig:multi-critic}. We define three reward groups, each associated with a different critic:
\begin{itemize}
    \item \textbf{Gliding Critic}: Responsible for evaluating rewards related to gliding, such as speed tracking.
    \item \textbf{Pushing Critic}: Focused on sparse contact-related rewards during the pushing phase.
    \item \textbf{Sim2Real Critic}: Responsible for rewards related to sim-to-real adaptation, such as action regularization.
\end{itemize}

Each critic is updated separately to estimate its value function. When calculating the overall advantage, the three critics' outputs are normalized and combined using weighted summation. Similar to \citet{zargarbashi2024robotkeyframing}, the weights for each critic’s advantage are treated as hyperparameters. This approach reduces the sensitivity to reward tuning, thereby simplifying the reward design process. Specifically, the reward function are presented in Table. \ref{tab:reward-group} and their values are in Appendix \ref{apx:mc_loss}.


\begin{table}[t]
\centering
\caption{Summary of Reward Terms and Their Expressions. 
Each term is multiplied by its phase coefficient 
(\(\delta_{\mathrm{glide}}\) or \(\delta_{\mathrm{push}}\)) 
if it belongs to a specific phase, and scaled by 
\(w_{\cdot}\) as listed in the text.}
\begin{tabular}{ll}
\toprule
\textbf{Gliding Critic Reward} & \textbf{Expression}\\
\midrule
\textit{Feet on board} 
& 
\(
\delta_{\mathrm{glide}}
\sum_{i=1}^{4} \!\Bigl(\|\mathbf{p}_{\mathrm{feet},i} - \mathbf{p}_{\mathrm{glide},i}\| < 0.05\Bigr)
\)
\\

\textit{Contact number} 
&
\(
\delta_{\mathrm{glide}} 
\ R_{contact\_num} \ref{eq:contact-num}
\)
\\

\textit{Feet distance} 
&
\(
\delta_{\mathrm{glide}}
\exp\!\Bigl(
-\,\sum_{i=1}^{4}
\|\mathbf{p}_{\mathrm{glide},j} - \mathbf{p}_{\mathrm{feet},j}\|
\Bigr)
\)
\\

\textit{Joint positions} 
&
\(
\delta_{\mathrm{glide}}
\exp\!\Bigl(
-\!\sum_{i=1}^{12}
(q_{i} - q^{\mathrm{glide}}_{i})^{2}
\Bigr)
\)
\\

\textit{Hip positions} 
&
\(
\delta_{\mathrm{glide}}
\exp\!\Bigl(
-\!\sum_{i \in {Hip}}
(q_{i} - q^{\mathrm{glide}}_{i})^{2}
\Bigr)
\)
\\
\midrule

\textbf{Pushing Critic Reward} & \textbf{Expression}\\
\midrule

\textit{Tracking linear velocity} 
&
\(
\delta_{\mathrm{push}}
\exp\!\Bigl(
-\tfrac{1}{\sigma}
\bigl\|
\mathbf{v}^{\mathrm{cmd}}_{x} - \mathbf{v}_{x}
\bigr\|^{2}
\Bigr)
\)
\\

\textit{Tracking angular velocity} 
&
\(
\delta_{\mathrm{push}}
\exp\!\Bigl(
-\tfrac{(\omega^{\mathrm{cmd}}_{z} - \omega_{z})^{2}}{\sigma_{\mathrm{yaw}}}
\Bigr)
\)
\\

\textit{Hip positions} 
&
\(
\delta_{\mathrm{push}}
\exp\!\Bigl(
-\!\sum_{i \in {Hip}}
(q_{i} - q^{\mathrm{push}}_{i})^{2}
\Bigr)
\)
\\

\textit{Orientation} 
&
\(
\delta_{\mathrm{push}}
\,\|\mathbf{g}_{xy}\|^{2}
\)
\\
\midrule

\textbf{Sim2Real Critic Reward} & \textbf{Expression}\\
\midrule

\textit{Wheel contact number} 
& 
\(
\;\!\Bigl(
\textstyle{\sum_{i \in \mathrm{wheels}} c_i} \;=\;4
\Bigr)
\)
\\

\textit{Board--body height} 
&
\(
\exp\!\Bigl(
-4\,\bigl|\,(z_{\mathrm{body}} - z_{\mathrm{board}}) \;-\; 0.15\bigr|
\Bigr)
\)
\\

\textit{Joint acceleration} 
&
\(
\sum_{i=1}^{12}
\biggl(
\frac{\mathrm{clip}\bigl(\dot{q}_{i}^{(t-1)} - \dot{q}_{i}^{(t)},\,{-10},\,10\bigr)}{\Delta t}
\biggr)^{2}
\)
\\

\textit{Collisions} 
&
\(
\sum_{i \in \mathcal{P}}
\!\Bigl(\|\mathbf{f}_{i}\| > 0.1\Bigr)
\)
\\

\textit{Action rate} 
&
\(
\,\bigl\|\mathbf{a}^{(t)} - \mathbf{a}^{(t-1)}\bigr\|
\)
\\

\textit{Delta torques} 
&
\(
\,\bigl\|\boldsymbol{\tau}^{(t)} - \boldsymbol{\tau}^{(t-1)}\bigr\|^{2}
\)
\\

\textit{Torques} 
&
\(
\,\|\boldsymbol{\tau}\|^{2}
\)
\\

\textit{Linear velocity (z-axis)} 
&
\(
\,\bigl\|\mathrm{clip}(v_{z},\,{-1.5},\,1.5)\bigr\|^{2}
\)
\\

\textit{Angular velocity (x/y)} 
&
\(
\bigl\|\mathrm{clip}(\boldsymbol{\omega}_{xy},\,{-1},\,1)\bigr\|^{2}
\)
\\

\textit{Base orientation} 
&
\(
\,\|\mathbf{g}_{xy}\|^{2}
\)
\\
\midrule

\textbf{Cycle Calculation} & \textbf{Expression}\\
\midrule

\textit{Cycle} & $T$\\
\textit{Phase} & $\phi \;\leftarrow\; \sin\!\bigl(2\pi\,t/T\bigr)$\\
\textit{Still Indicator} & $\delta_{still}$\\
\textit{Glide Indicator} & 
$
\delta_{\mathrm{glide}}^{(t)} 
\;=\; 
LPF\Bigl(\bigl[\phi < 0.5\bigr] \;\lor\; \delta_{\mathrm{still}}\Bigr)
$\\

\textit{Push Indicator} &

$
\delta_{\mathrm{push}}^{(t)} 
\;=\; 
LPF\Bigl(\bigl[\phi \ge 0.5\bigr] \;\land\; \neg\,\delta_{\mathrm{still}}\Bigr)
$\\

\textit{Low pass filter} & $LPF$\\

\midrule

\bottomrule
\end{tabular}

\label{tab:reward-group}
\end{table}

\subsection{Beta Distribution Policy}
\label{beta}

In quadrupedal locomotion tasks, the mainstream frameworks typically assume a Gaussian distribution as the policy distribution in PPO due to its intuitive parameterization and ease of shape control \cite{rudin2022learning}. However, when the action space has strict bounds (e.g., joint position limits to prevent collisions in quadrupedal robots), the Gaussian distribution can introduce bias in policy optimization by producing out-of-bound actions that need to be clipped \cite{chou2017improving}. While this issue has been noted in prior work, it has not been widely addressed in the context of robotic locomotion.

We observe that the Beta distribution offers a significant advantage over the Gaussian distribution in effectively utilizing the action space under sparse reward conditions. In contrast, Gaussian policies may increase the variance excessively in an attempt to explore more of the action space to trigger potential sparse rewards. This aggressive strategy can lead to suboptimal performance, getting stuck in local optima, and even result in hardware damage and safety hazards during real-world deployment. Therefore, we introduce the Beta distribution as the policy distribution for our framework.

In our implementation, the policy outputs the shape parameters \((\alpha, \beta)\) of the Beta distribution for each joint independently. To ensure that the Beta distribution remains unimodal (\(\alpha > 1, \beta > 1\)), we modify the activation function as follows:

\begin{equation} 
\text{SoftplusWithOffset}(x) = \log(1 + e^x) + 1 + 10^{-6} 
\end{equation}

The standard Beta distribution is defined over \([0, 1]\):

\begin{equation}
f(x; \alpha, \beta) = \frac{\Gamma(\alpha + \beta)}{\Gamma(\alpha) \Gamma(\beta)} x^{\alpha - 1} (1 - x)^{\beta - 1}, \quad \alpha, \beta > 0
\end{equation}

To adapt this to the robot's action space \([ -a_{\text{max}}, a_{\text{max}} ]\) where \( a_{\text{max}} > 0 \), we apply the following transformation:

\begin{equation}
a \sim \mathcal{B}(\alpha, \beta), \quad a' = a \cdot 2a_{\text{max}} - a_{\text{max}}
\end{equation}

In the expression, the $a'$ denote the true action will be executed by robot. This approach enables the policy to produce bounded, valid actions within the desired range, preventing out-of-bound behaviors while making full use of the action space. We provide a proof in the Appendix \ref{app:beta} demonstrating that our method does not introduce bias or result in variance explosion.

\section{EXPERIMENTS}

In this section, the purpose of our experiment is to verify whether our method can achieve the following behaviors:
\begin{itemize}
    \item \textbf{Q1: } Why Hybrid Dynamics System is better than Single Dynamics modeling?
    \item \textbf{Q2: } Can our method identify the mode of skateboarding?
    \item \textbf{Q3: } Can our method achieve skateboarding in the real-world with disturbances?
\end{itemize}

\subsection{Prediction of Dynamics}

\begin{figure}
    \centering
    \includegraphics[width=1\linewidth]{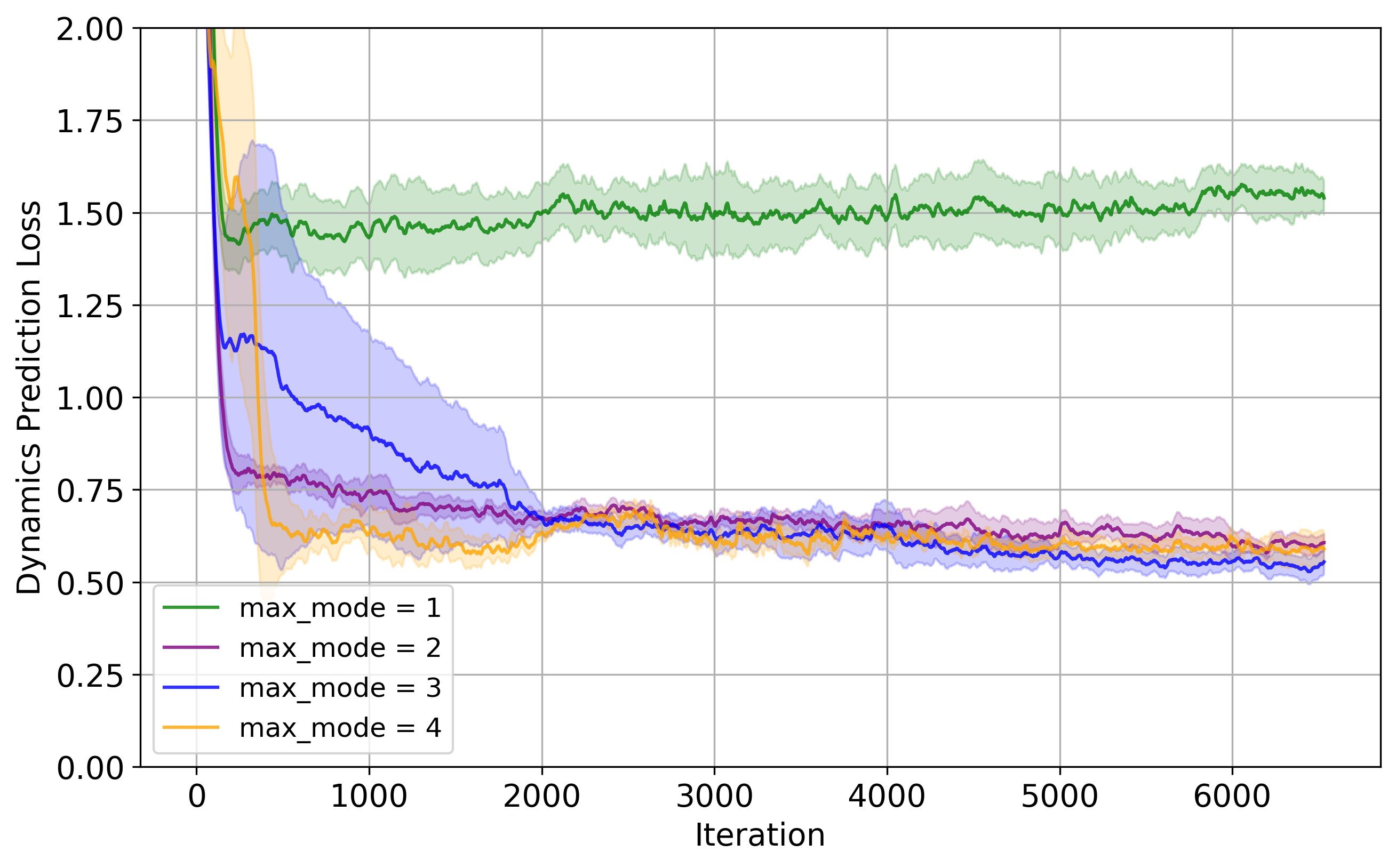}
    \caption{Dynamics Prediction Loss: The agent’s dynamics prediction loss $\texttt{MSE}(\hat{o}_{t+1},{o}_{t+1})$ during training is shown in the figure, where the thick line represents the average loss, and the shaded regions indicate the confidence intervals across different seeds.}
    \label{fig:dynamics-loss}
\end{figure}

To answer \textbf{Q1}, We apply comparison on dynamics prediction loss between different maximal modes, corresponding to dimension of mode indicator $|\delta|$. Among these modes, the condition where number of mode $|\delta|$ equals to 1 represents that using one network to model whole dynamics like \cite{10161144}. We trained each condition three times with random seeds for network initialization. As shown in Fig. \ref{fig:dynamics-loss}, the curve shows the average reconstruction loss under different modes. When the maximum number of modes is 1, the loss is the highest. After incorporating the hybrid dynamics idea, different modes are switched to guide the conversion and mutation of flow dynamics, the reconstruction loss is smaller. Starting from the maximum number of modes 2, the improvement in prediction accuracy begins to plateau.

This is consistent with our assumption that it is more reasonable to build the system as a hybrid dynamics system in a system with mutation and other properties. Considering that the two main states of the skateboard are the skateboard up and the skateboard down, when $|\delta| \geq 2$, there is a marginal effect of improvement. When $|\delta| \geq 4$, the prediction accuracy can hardly be improved and converges to the state of mode=3. \textbf{It is worth noting that we only set the maximum number of modes to 3, rather than requiring that all 3 modes must be present.} Therefore, in subsequent experiments, we believe that a mode count of 3 is the reasonable maximal number of modes for this system, which corresponds to three motion modes: on the skateboard, pushing the skateboard under the skateboard, and being in the air between the two. We will showcase the mode identification in the next section \ref{sec:mode_identification}.

\begin{figure}
    \centering
    \includegraphics[width=1\linewidth]{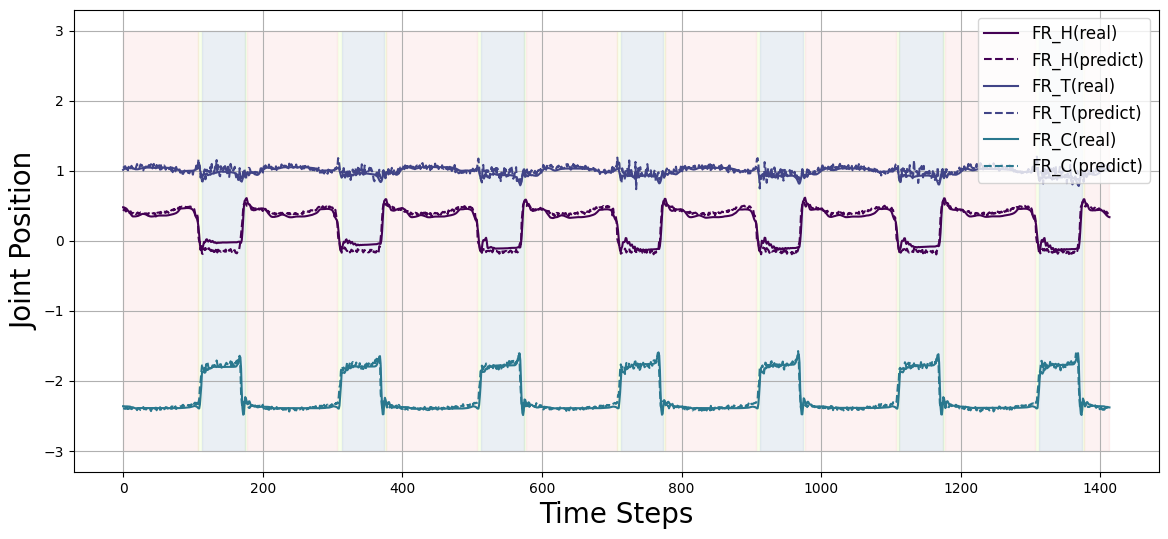}
    \caption{Trajectory Prediction Visualization: The comparison between the actual position trajectory (solid line) and the predicted position trajectory (dashed line) for the right front leg joint motor of the physical Go1 robot during skateboarding is shown. We selected the right front leg joint, which exhibits the largest range of motion, as the visualization target. During deployment, the system utilizes the mode selection results from the automata to choose the corresponding decoder for prediction, consistent with the training process.}
    \label{fig:joint-pred}
\end{figure}

To validate the accuracy of the predicted dynamics, we deployed our DHAL algorithm on the physical robot and recorded both the predicted dynamics and the actual dynamics in real time. In this experiment, we applied a clock signal with a fixed period of 2.5 seconds to the controller, alternating the skateboard's upward and downward movements. Notably, this clock signal differs from the one used during the training phase (4s), providing an additional test for the accuracy of the predicted dynamics under new conditions. As shown in Fig. \ref{fig:joint-pred}, even in the absence of $u_t$, the representation obtained only from $x_{t-k:t}$ can accurately predict the robot's trajectory and jumps. In the next section, we will evaluate our method's capability for mode identification and its ability to adjust phases on the fly.


\subsection{Mode Identification}
\label{sec:mode_identification}

\begin{figure*}[]
    \centering
    \includegraphics[width=1\linewidth]{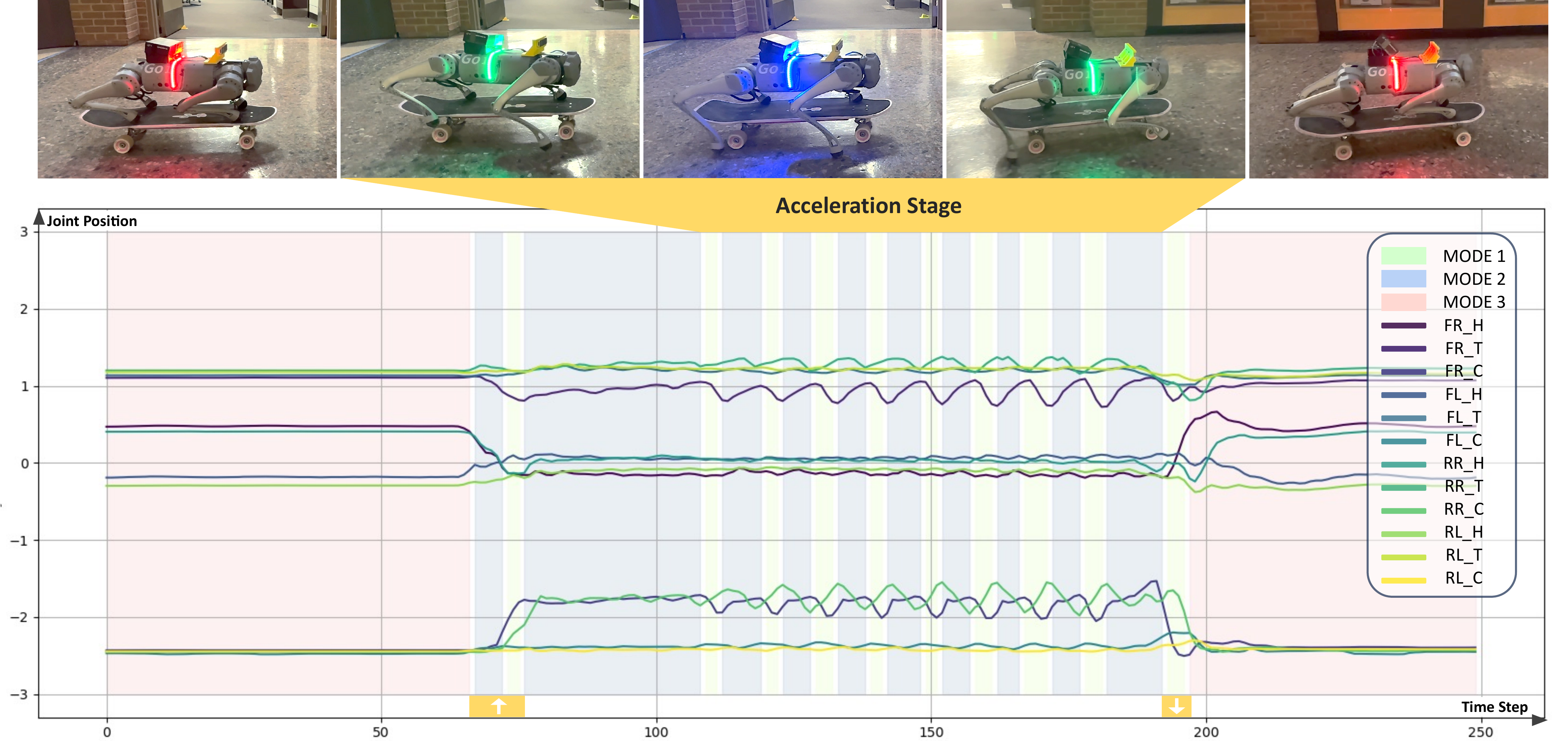}
    \caption{Effectiveness of mode identification. In real-world deployment, we light up different RGB light bar colors according to the mode to show the switching between different mode. The following figure shows the change in joint position relative to time in the test, and the background color is represented by the color of the corresponding mode. [H, T, C] denote the Hip, Thigh, and Calf Joints, respectively.} 
    \label{fig:model-id}
\end{figure*}

\begin{figure}[]
    \centering
    \includegraphics[width=1\linewidth]{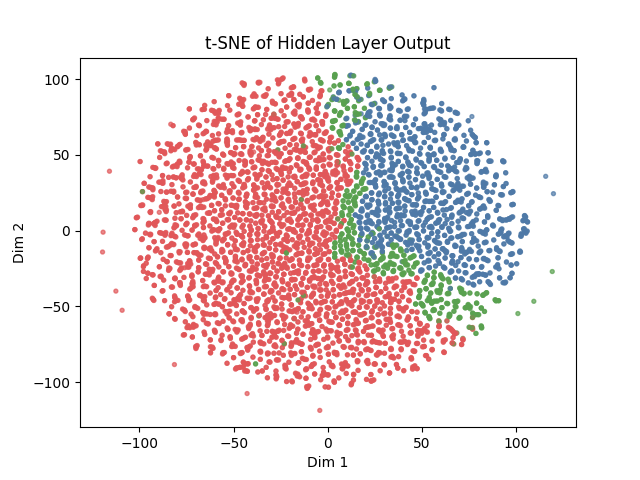}
    \caption{Visualization of hidden layer of the controller: Scatter points in different colors correspond to the different modes identified by the system, consistent with Fig. \ref{fig:model-id}. Specifically, [green, blue, red] represent [mode 1, mode 2, mode 3], respectively.}
    \label{fig:tsne}
\end{figure}

To answer \textbf{Q2}, we collected real-world trajectories of the robot skateboarding along with the modes selected by the controller for visualization, as shown in Fig. \ref{fig:model-id}. Additionally, we used different RGB LED colors to represent the dynamic modules selected by the hybrid automata. From the figure, it is visually evident that when the robot is in the gliding phase, with all four feet standing firmly on the skateboard, the hybrid automata classify this state as mode 3 (red). When the robot's right front foot starts to lift off the ground and enters the swing phase, the automata naturally transition the mode to mode 1 (green). This brief transition corresponds to the sharp change in joint angles shown in the figure. Once both of the robot's right legs are fully in contact with the ground, the automata switch to mode 2 (blue).

This sequence of mode selections and transitions is smooth and explicitly aligns with the decomposition of skateboarding motion: (1) gliding phase, (2) airborne phase transitioning on and off the skateboard, and (3) pushing phase. These results demonstrate that our hybrid automata make mode selections that are highly consistent with physical intuition.

Additionally, we applied t-distributed stochastic neighbor embedding (t-SNE) to reduce the dimensionality of the hidden layer outputs from the controller's neural network. As illustrated in Fig. \ref{fig:tsne}, the latent space exhibits a clearly defined distribution across different modes. Interestingly, the resulting latent structure is remarkably similar to that reported in \cite{10197536}. Specifically, the red region primarily corresponds to the data collected during the robot's movements on the skateboard, the blue region represents states where the right two legs are in contact with the ground during pushing, and the green region corresponds to the airborne phases when transitioning on and off the skateboard.

This observation highlights two key points: (1) our controller effectively handles motion control tasks across different modes, and (2) our DHAL module can distinctly and accurately differentiate between various modes.

\subsection{Performance on Skateboarding}

\begin{figure}
    \centering
    \includegraphics[width=1\linewidth]{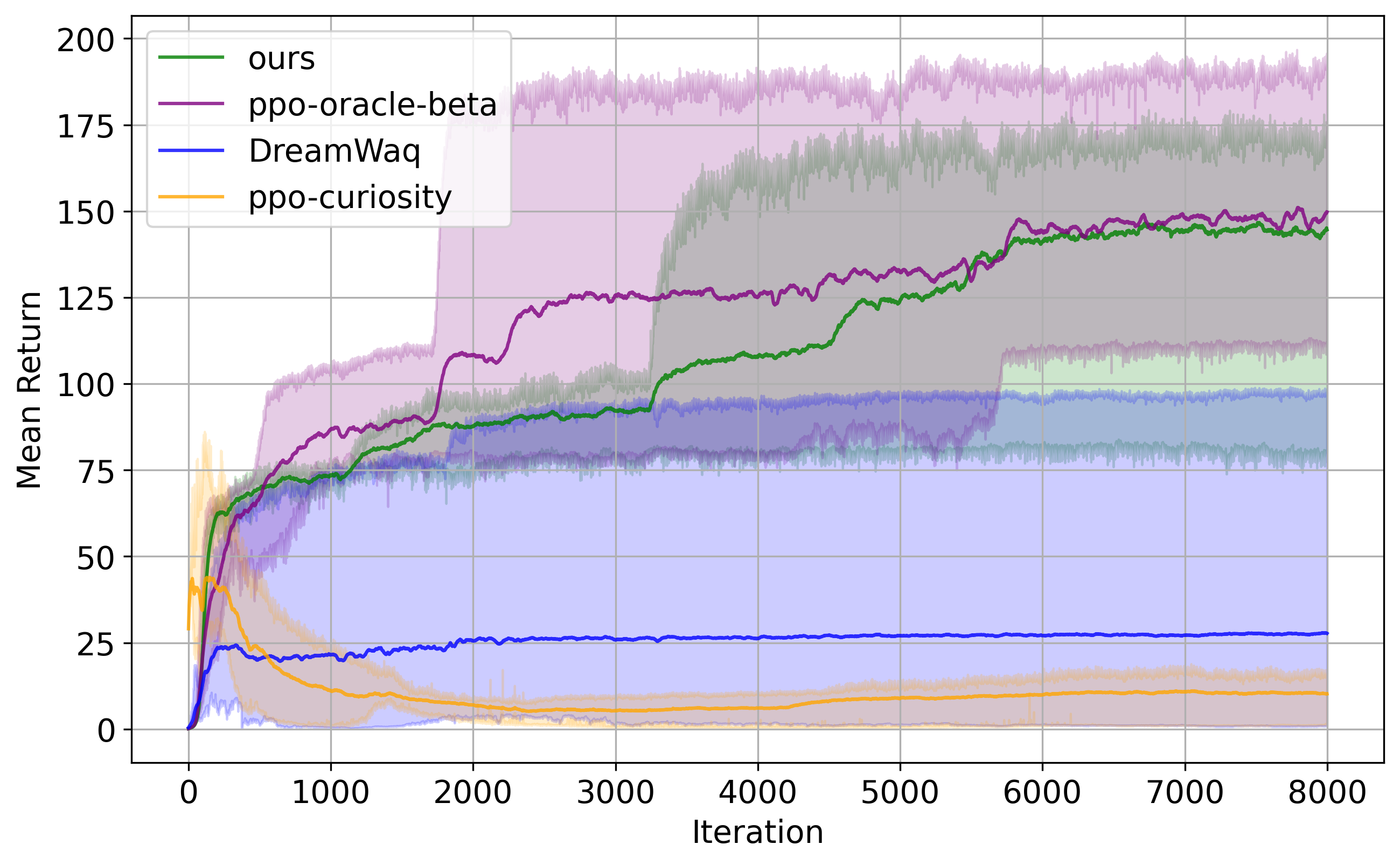}
    \caption{Comparison of Training Rewards: Comparison of mean reward during training is shown in the figure, where the thick line represents the average return, and the shaded regions indicate the maximal and minimal reward across different seeds. Each method was trained using four random seeds to evaluate performance.}
    \label{fig:reward}
\end{figure}

To answer \textbf{Q3}, this section presents ablation studies and real-world experiments to evaluate our method. The analysis is divided into the following parts: (1) Comparison of training returns, (2) Comparison between single-critic method, and (3) Experiments in real-world scenarios with disturbances, including success rate statistics.

\textbf{Comparison of training returns:} Since the skateboarding task is designed to be contact-guided, the training process exhibits significant randomness, leading to considerable variance in training curves even for the same method. Therefore, the primary goals of this experiment are: (1) to identify the key factors driving successful training, and (2) to evaluate whether our method can approach optimal performance. 

For a comparative evaluation, we compared the following algorithms with access to proprioceptions only:

\begin{enumerate}
    \item \textbf{PPO-oracle-beta}: Training a policy with full privileged observations and the Beta distribution. 
    \item \textbf{DreamWaq}~\cite{10161144}: Training an dynamics module to estimate velocity and future observation. 
    \item \textbf{PPO-curiosity}~\cite{rudin2022learning, zhang2024wococo}: Training directly with only proprioception and following the curiosity reward design\cite{zhang2024wococo}.
\end{enumerate}

As shown in Fig. \ref{fig:reward}, our method could achieve comparable performance with \textbf{PPO-oracle-beta}, which has privileged observation about skateboard information. Notably, for the methods with Gaussian distribution, the robot cannot learn how to do skateboarding even leading to dangerous motion, which makes real-world deployment infeasible. The result aligns with the discussion in Section \ref{beta}: in environments with high exploration difficulty, Gaussian distributions tend to prioritize increasing variance to expand the exploration range, thereby randomly encountering ``reward points''. However, due to the physical constraints of the robot, this exploration strategy introduces bias, causing Gaussian distribution policies to favor movements closer to the constraints and ultimately leading to failure \cite{chou2017improving}.

\textbf{Comparison with single-critic:} We trained both our multi-critic method and the single-critic method for comparison. Since our multi-critic approach normalizes the advantages of different reward groups and combines them through weighted summation, while the single-critic approach lacks this weighting advantage mechanism, we evaluated two configurations of the single-critic method:

\begin{itemize}
    \item \textbf{Single-critic-w-transfer}: A single-critic setup with the same reward configuration as the multi-critic method, but with new reward weights transferred based on the advantage weights
    \item \textbf{Single-critic-wo-transfer}: A single-critic setup with the same reward configuration and weights as the multi-critic method
\end{itemize}

\begin{figure}
    \centering
    \includegraphics[width=1\linewidth]{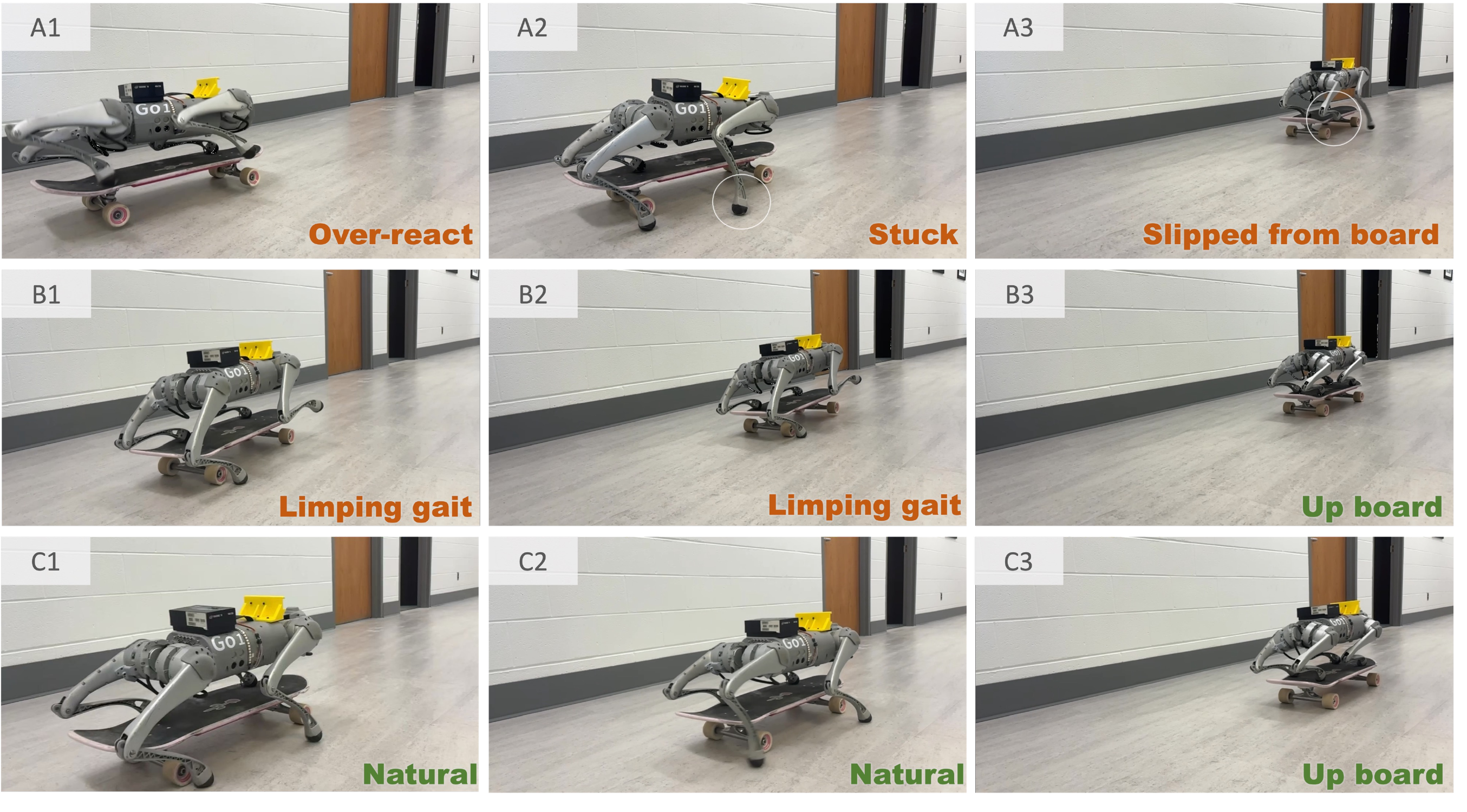}
    \caption{Comparison between single-critic policy and multi-critic policy: Single-critic-wo-transfer ($A1\sim A3$), Single-critic-w-transfer ($B1\sim B3$), ours ($C1\sim C3$).}
    \label{fig:comparison-critic}
\end{figure}

As shown in Fig. \ref{fig:comparison-critic}, For the single-critic approach without transferred weights, the robot exhibited aggressive and erratic movements, making it difficult to handle disturbances. During forward motion, excessive hyperflexion at the foot caused it to get stuck, and when mounting the skateboard, the hind legs often slipped off. In contrast, the single-critic approach with transferred weights managed to successfully mount the skateboard. However, during the pushing phase, the robot primarily relied on its hind legs, leaving the front legs suspended for extended periods, resulting in an unnatural gliding posture. 

With our multi-critic training scheme, the robot achieved a smooth and natural motion, efficiently executing rapid pushing and demonstrating significantly more stable and graceful transitions on and off the skateboard. We obtained results similar to \citet{mysore2022multi}, demonstrating that the multi-critic approach is well-suited for multi-style learning.

\textbf{Quantitative Experiments:} We conducted quantitative experiments in real-world scenarios to evaluate whether our method can successfully complete the skateboarding task under real-world noise and disturbances. These disturbances include, but are not limited to, sensor noise, skateboard property variations, terrain irregularities, and dynamics noise \cite{gu2024advancing}. Using the trained model, we tested the following scenarios: (1) smooth ceramic flooring, (2) soft carpeted flooring, (3) disturbance, (4) slope terrain, (5) single-step terrain, (6) uneven terrain. Each scenario was tested ten times, with each test containing at least one full cycle of mounting and dismounting the skateboard. The test terrain is shown in Fig. \ref{fig:cover} and success rate statistics are shown in Table. \ref{tab:quantitative experiment}. More extreme terrain experiments and validation in other task could be found in Appendix \ref{app:exp}.

\begin{table}
    \caption{Succes Rate Comparison: We deployed each method on a real robot to evaluate the success rate. Each method was tested five times per scenarios. Success was defined as completing at least one full up-board and down-board motions, traverse over a distance of more than 5 meters, and avoiding abrupt movements or detachment from the skateboard. \ding{55} indicates complete failure(Massive torque caused the joints protection state, for hardware protection, we first test torque value in simulation to make sure it will no exceed safty range).}
    \centering
    \begin{tabular}{cccc}
    \toprule
        Method & \textbf{Ceramic} & \textbf{Carpet} & \textbf{Disturbance} \\
        \midrule
        \textbf{Ours} & \textbf{100\%} & \textbf{100\%} & \textbf{100\%}\\
        \textbf{Our-wo-MC(transfered)} & 100\% & 100\% & 60\%\\
        \textbf{Our-wo-MC} & 60\% & 60\% & 40\% \\
        \textbf{Ours-wo-Beta} & \ding{55} & \ding{55} & \ding{55}\\
        \textbf{DreamWaq\cite{10161144}} & \ding{55} & \ding{55} & \ding{55}\\
        
        \midrule
        Method & \textbf{Slope} & \textbf{Single-step} & \textbf{Uneven}\\
        \midrule
        \textbf{Ours} & \textbf{80\%} & \textbf{100\%} & \textbf{60\%}\\ 
        \textbf{Our-wo-MC(transfered)} & 60\% & 40\% & 60\%\\
        \textbf{Our-wo-MC} & 0\% & 40\% & 40\%\\
        \textbf{Ours-wo-Beta} & \ding{55} & \ding{55} & \ding{55}\\
        \textbf{DreamWaq\cite{10161144}} & \ding{55} & \ding{55} & \ding{55}\\
            
        \bottomrule
    \end{tabular}

    \label{tab:quantitative experiment}
\end{table}

%
\section{Limitations and discussion} 
\label{sec:conclusion}

1. \textbf{Perception Limitations}: To connect the robot’s left front foot to the skateboard(only one foot), we assumed a spherical joint to prevent the skateboard from completely detaching from the robot. The transition from walking to skateboarding presents a significantly greater challenge, requiring hardware modifications, such as adjusting the camera layout and incorporating multiple cameras to locate the skateboard. Furthermore, we did not consider obstacle avoidance during skateboarding using perception-based methods. Initially, we attempted to use the Realsense T265 for state estimation but later determined that it was unnecessary for this task. However, for future work, when the foot is not fixed to the board, the state estimation methods \cite{teng2021legged, yu2023fully, he2024legged, teng2024gmkf, ghaffari2022progress} need to be carefully integrated. 

2. \textbf{Complex skill Generalization}: Our method cannot generalize to extreme skateboarding techniques equivalent to those of human athletes, such as performing an ollie. The current simulation setup cannot accurately replicate the motion and contact dynamics of passive wheels in such challenging scenarios. Instead, we relied on approximations and alternative techniques to simulate these dynamics as realistically as possible.

3. \textbf{Limitations in Dynamics Learning}: The learned dynamics are not yet precise enough for model-based control. Furthermore, the coupling between the controller and the dynamics predictor prevents iterative optimization, such as that used in MPC, limiting the flexibility and efficiency of our approach.

4. \textbf{Non-Trivial Environment Design:} The environment setting and design for robot skateboarding in non-trivial. This part requires manual design and inspection. We believe that in the future, integrating environment generation with large models~\cite{Genesis} could potentially help address this challenge.
\section{Conclusion} 
\label{sec:conclusion}

We proposed the Discrete-time Hybrid Automata Learning (DHAL) framework to address mode-switching in hybrid dynamical systems without requiring trajectory segmentation or event function modeling. By combining a multi-critic architecture and a Beta distribution policy, our method demonstrates robust handling of contact-guided hybrid dynamics, as validated through the challenging task of quadrupedal robot skateboarding. Real-world experiments showed that our approach achieves smooth and intuitive mode transitions, effectively balancing gliding and pushing behaviors. While limitations remain, such as terrain generalization and coupling between the controller and dynamics predictor, DHAL offers a promising step toward learning-based control for hybrid systems in robotics.

\section*{Acknowledgments}
M. Ghaffari was supported by AFOSR MURI FA9550-23-1-0400. We appreciate the valuable discussions, hardware guidance and constructive feedback from Yulun Zhuang and Yi Cheng. We also extend our gratitude to Linqi Ye for the initial brainstorming and insightful suggestions.

{\small
\balance
\bibliographystyle{unsrtnat}
\bibliography{references}
}

\begin{appendices}
\section{Multi-Critic}
\label{apx:mc_loss}
\subsection{Multi-Critic PPO Loss design}

\begin{figure*}
    \centering
    \includegraphics[width=1\linewidth]{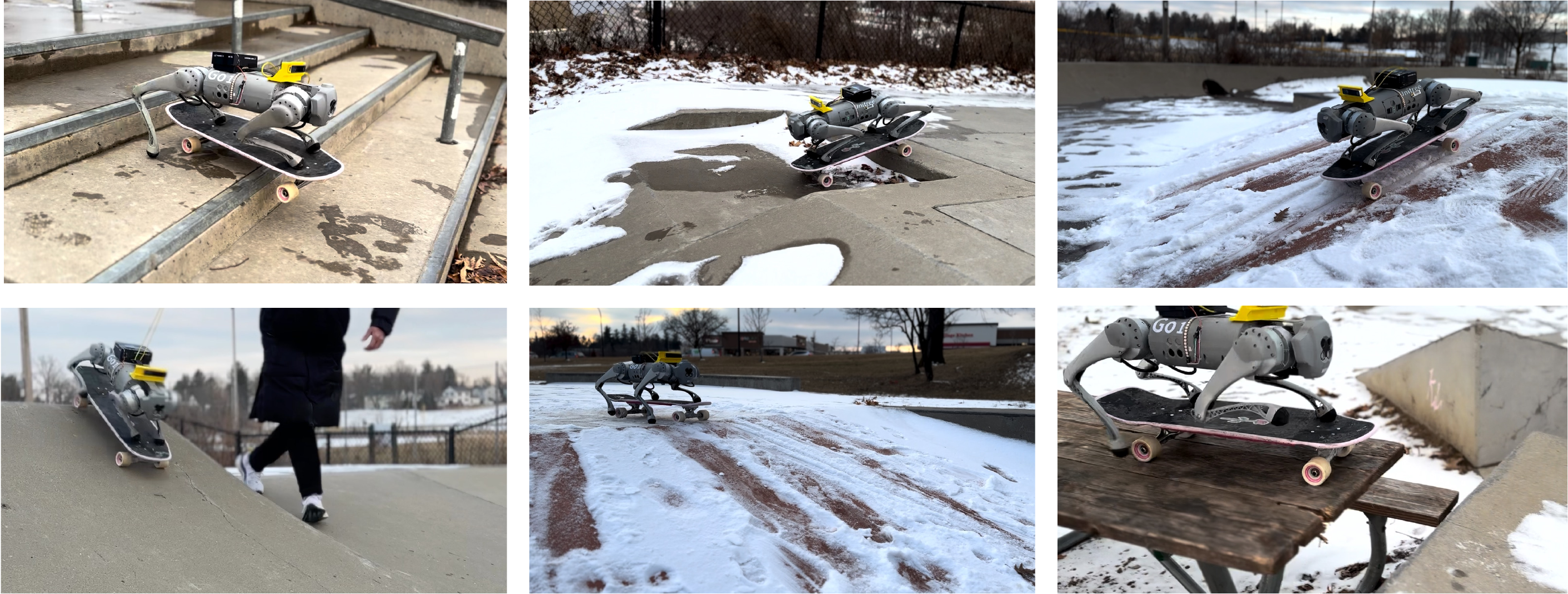}
    \caption{Real-world Experiments in Skateboard Park. For additional
demonstrations, please refer to the our \href{https://umich-curly.github.io/DHAL/}{website},
where more result videos are available.} 
    \label{fig:skbp}
\end{figure*}

Motivated by \citet{zargarbashi2024robotkeyframing}, we define \(\mathcal{P}\), \(\mathcal{G}\), and \(\mathcal{S}\) to represent the "Pushing," "Gliding," and "Sim2Real" tasks, respectively. Consequently, \(r_{\mathcal{P}}, r_{\mathcal{G}}, r_{\mathcal{S}}\) denote the weighted sums of the specific reward groups, while \(V_{\mathcal{P}}, V_{\mathcal{G}}, V_{\mathcal{S}}\) correspond to the respective value networks. The overall value loss is given by \(L^{\text{value}} = L_{\mathcal{P}} + L_{\mathcal{G}} + L_{\mathcal{S}}\), where each term is defined as:

\begin{equation}
L_{\mathcal{P}} = \mathbb{E}_t \left[ \left\| r_{{\mathcal{P}},t} + \gamma V_{\mathcal{P}}(s_{t+1}) - V_{\mathcal{P}}(s_t) \right\|^2 \right]
\end{equation}
\begin{equation}
L_{\mathcal{G}} = \mathbb{E}_t \left[ \left\| r_{{\mathcal{G}},t} + \gamma V_{\mathcal{G}}(s_{t+1}) - V_{\mathcal{G}}(s_t) \right\|^2 \right]
\end{equation}
\begin{equation}
L_{\mathcal{S}} = \mathbb{E}_t \left[ \left\| r_{{\mathcal{S}},t} + \gamma V_{\mathcal{S}}(s_{t+1}) - V_{\mathcal{S}}(s_t) \right\|^2 \right]
\end{equation}

Here, \(\gamma\) is the discount factor, and \(s_t\) represents the state at time \(t\). Each value loss minimizes the temporal difference (TD) error for the corresponding reward group.

For advantage estimation in PPO, each reward group and its associated critic calculate the advantage separately based on the TD error. Taking "Pushing" as an example, the TD error is defined as:

\begin{equation}
\delta_{\mathcal{P},t} = r_{\mathcal{P},t} + \gamma (1 - d_t) V_{\mathcal{P},t+1} - V_{\mathcal{P},t}
\end{equation}

where \(d_t\) is an indicator variable denoting whether the episode terminates at time \(t\). The advantage is then calculated recursively as:

\begin{equation}
A_{\mathcal{P},t} = \delta_{\mathcal{P},t} + \gamma (1 - d_t) \lambda A_{\mathcal{P},t+1}
\end{equation}

Here, \(\lambda\) is the Generalized Advantage Estimation (GAE) parameter, which balances the trade-off between bias and variance in advantage estimation. After calculating the advantage for "Pushing," it is normalized as follows:

\begin{equation}
\tilde{A}_{\mathcal{P},t} = \frac{A_{\mathcal{P},t} - \mu_{A_{\mathcal{P},t}}}{\sigma_{A_{\mathcal{P},t}} + \epsilon}
\end{equation}

where \(\mu_{A_{\mathcal{P},t}}\) and \(\sigma_{A_{\mathcal{P},t}}\) are the mean and standard deviation of the advantage values for the "Pushing" task, and \(\epsilon\) is a small constant added for numerical stability. This process is repeated for both "Gliding" and "Sim2Real," resulting in normalized advantages \(\tilde{A}_{\mathcal{G},t}\) and \(\tilde{A}_{\mathcal{S},t}\). Finally, the weighted sum of all normalized advantages is computed as:

\begin{equation}
\tilde{A}_t = w_1 *\tilde{A}_{\mathcal{P},t} + w_2 * \tilde{A}_{\mathcal{G},t} + w_3 * \tilde{A}_{\mathcal{S},t}
\end{equation}

The surrogate loss is then calculated as in the standard PPO process:

\begin{equation}
L^{\text{surrogate}} = \mathbb{E}_t \left[ \min \left( \alpha_t \tilde{A}_t, \, \text{clip}(\alpha_t, 1 - \epsilon, 1 + \epsilon) \tilde{A}_t \right) \right]
\end{equation}

where \(\alpha_t\) is the ratio between the new policy and the old policy probabilities. Finally, the overall PPO loss is computed as:

\begin{equation}
L^{\text{PPO}} = L^{\text{value}} + L^{\text{surrogate}} - c \cdot \mathcal{H}(\pi_\theta)
\end{equation}

Here, \(\mathcal{H}(\pi_\theta)\) represents the entropy of the policy \(\pi_\theta\), encouraging exploration, and \(c\) is a weighting coefficient. This formulation integrates the value loss, the surrogate loss, and an entropy regularization term to achieve robust and efficient policy optimization.

\subsection{Reward Detail}

The contact number reward for the gliding phase is expressed as:

\begin{equation} \label{eq:contact-num}
    R = 2 \cdot R_{skateb} - P_{ground}
\end{equation}

where \( R_{skateb} \) represents the reward for maintaining correct contact with the skateboard, and \( P_{ground} \) is the penalty for undesired ground contact.

The skateboard contact reward is given by:

\begin{equation}
R_{skateb} = \delta_{glide} \cdot ( \sum_{i=1}^{4} \mathds{1}(c_{skateb,i}) + 4 \cdot \mathds{1}(\sum_{i=1}^{4} \mathds{1}(c_{skateb,i}) = 4) ),
\end{equation}

where \( \delta_{glide} \) is a scaling coefficient for the gliding phase, \( \mathds{1}(c_{skateb,i}) \) indicates whether the \(i\)-th foot is in contact with the skateboard, and an additional reward is provided if all four feet maintain contact.

The ground contact penalty is expressed as:

\begin{equation}
P_{ground} = \delta_{glide} \cdot ( \sum_{i \in \{0,2\}} \mathds{1}(\sim c_{skateb,i}) + \sum_{i \in \{0,2\}} \mathds{1}(c_{ground,i})),
\end{equation}

where \( \mathds{1}(\sim c_{skateb,i}) \) penalizes the lack of skateboard contact for specific feet, and \( \mathds{1}(c_{ground,i}) \) penalizes unintended ground contact.

This reward design encourages the agent to maintain stable contact with the skateboard while avoiding unnecessary ground contact, ensuring smooth and efficient gliding behavior.

The detailed reward weights is shown in Table. \ref{tab:reward-group-weight}

\begin{table}[t]
\caption{Reward weights and Advantage weights for skateboarding environment design}
\centering
\begin{tabular}{ll}
\toprule
\textbf{Gliding Critic Reward} & \textbf{Weight}\\
\midrule
\textit{Feet on board} & 0.3\\
\textit{Contact number} & 0.3\\
\textit{Feet distance}  & 1.8\\
\textit{Joint positions}  & 1.2\\
\textit{Hip positions}  & 1.2\\
\midrule

\textbf{Pushing Critic Reward} & \textbf{Weight}\\
\midrule
\textit{Tracking linear velocity}  & 1.6\\
\textit{Tracking angular velocity}  & 0.8\\
\textit{Hip positions}  & 0.6\\
\textit{Orientation}  & -2\\
\midrule

\textbf{Sim2Real Critic Reward} & \textbf{Weight}\\
\midrule
\textit{Wheel contact number}  & 0.8\\
\textit{Board--body height}  & 1\\
\textit{Joint acceleration}  & -2.5e-7\\
\textit{Collisions}  & -1\\
\textit{Action rate}  & -0.22\\
\textit{Delta torques}  & -1.0e-7\\
\textit{Torques}  & -1.0e-5\\
\textit{Linear velocity (z-axis)}  & -0.1\\
\textit{Angular velocity (x/y)}  & -0.01\\
\textit{Base orientation}  & -25\\
\midrule

\textbf{Advantage} & \textbf{Weight}\\
\midrule
\textit{Gliding Critic Advantage}  & 0.35\\
\textit{Pushing Critic Advantage}  & 0.4\\
\textit{Sim2Real Critic Advantage}  & 0.25\\

\bottomrule
\end{tabular}

\label{tab:reward-group-weight}
\end{table}

\section{Beta Distribution}
\label{app:beta}

\subsection{Why Gaussian distribution policy introduce bias}

Based on \citet{chou2017improving}, when a Gaussian policy $\pi_\theta(a \mid s) = \mathcal{N}(\mu_\theta, \sigma_\theta^2)$ is employed in a bounded action space $[-h,h]$(whatever because of environment manually design or physical constraints of robots), any action $a$ exceeding these limits is clipped to $\mathrm{clip}(a)\in[-h,h]$. Ideally, the policy gradient should be computed as
\[
\nabla_\theta J(\pi_\theta)
=\mathbb{E}_{a\sim \pi_\theta}\bigl[\nabla_\theta \log \pi_\theta(a \mid s)\,A^\pi(s,a)\bigr].
\]
However, if the update actually uses the clipped action in the value function,
\[
g_{\text{clip}} 
= \nabla_\theta \log \pi_\theta(\mathrm{clip}(a) \mid s)\;A^\pi\bigl(s,\mathrm{clip}(a)\bigr),
\]
then integrating over all $a$ reveals a discrepancy whenever $\lvert a\rvert > h$. Specifically,

\[
\begin{array}{l} 
\mathbb{E}[g_{\text{clip}}] - \nabla_\theta J(\pi_\theta) = \\
\mathbb{E}_s \Bigg[ \int_{\lvert a \rvert > h} \pi_\theta(a|s) ( \nabla_\theta \log \pi_\theta(\pm h|s) A^\pi(s, \pm h)\\ - \nabla_\theta \log \pi_\theta(a|s) A^\pi(s, a) ) \, da \Bigg] \neq 0.
\end{array}
\]

Because a Gaussian often places nonnegligible probability mass outside $\pm h$, this mismatch fails to cancel out, producing a biased gradient that nudges the policy to favor actions beyond the valid range. The Policy Gradient of Gaussian distribution policy is shown below.

\[
\nabla_{\sigma_\theta} J(\theta) = \frac{1}{\sigma_\theta^3} \mathbb{E}_{a \sim \pi_\theta} \left[ \left( (a - \mu_\theta)^2 - \sigma_\theta^2 \right) A^\pi(s, a) \right].
\]

Moreover, due to the bias in the policy, it may tend to produce actions outside the valid range. Increasing the variance becomes a direct consequence of this tendency. This forms a positive feedback loop: the more the policy variance grows, the more out-of-bound actions are sampled, and the larger \((a-\mu_\theta)^2\) becomes in the gradient—even though the actions are physically clipped. Unlike the Beta policy, a Gaussian distribution can increase \(\mu\) indefinitely. That's the reason why Gaussian policy need more cautious reward design.

\subsection{Why Beta distribution do not introduce bias}

For Beta distribution, we first need to  rescale it from $[0,1]$ to $[-h,+h]$. Suppose $\tilde{z} \sim \text{Beta}(\alpha_\theta(s), \beta_\theta(s))$, which is supported on $[0, 1]$ and define $a = \phi(\tilde{z}) = 2h(\tilde{z} - \frac{1}{2})$. Thus $a \in [-h, h]$.

Because $\phi$ is a smooth, bijective function from $[0, 1] \to [-h, h]$, we can define the policy's pdf as:
\[
\pi_\theta(a \mid s) = \text{Beta}_{\alpha_\theta(s), \beta_\theta(s)}\left(\phi^{-1}(a)\right) \times \left| \frac{d}{da} \phi^{-1}(a) \right|.
\]

Concretely,
\[
\phi^{-1}(a) = \frac{a + h}{2h}, \quad \left| \frac{d}{da} \phi^{-1}(a) \right| = \frac{1}{2h}.
\]

Hence,
\[
\pi_\theta(a \mid s) = \text{Beta}_{\alpha_\theta, \beta_\theta}\left(\frac{a + h}{2h}\right) \times \frac{1}{2h}, \quad a \in [-h, h].
\]

Let us verify the critical zero-integral property for the Beta policy:
\[
\int_{-h}^{+h} \pi_\theta(a \mid s) \nabla_\theta \log \pi_\theta(a \mid s) \, da = \int_{-h}^{+h} \nabla_\theta \pi_\theta(a \mid s) \, da.
\]

But
\[
\int_{-h}^{+h} \pi_\theta(a \mid s) \, da = 1 \quad \text{(all the mass is inside } [-h, h]).
\]

Thus,
\[
\nabla_\theta \int_{-h}^{+h} \pi_\theta(a \mid s) \, da = \nabla_\theta [1] = 0.
\]

Hence,
\[
\int_{-h}^{+h} \nabla_\theta \pi_\theta(a \mid s) \, da = 0, \text{i.e.,} \int_{-h}^{+h} \pi_\theta(a \mid s) \nabla_\theta \log \pi_\theta(a \mid s) \, da = 0. 
\]

No boundary terms appear, because $\pi_\theta(a \mid s)$ is zero outside $[-h, h]$.Thus, if plug $\pi_\theta$ from above into the standard policy gradient formula (1), we could get an unbiased estimator:
\[
\mathbb{E} \big[ \nabla_\theta \log \pi_\theta(a \mid s) Q^\pi(s, a) \big] = \nabla_\theta \int_{-h}^{+h} \pi_\theta(a \mid s) Q^\pi(s, a) \, da 
\]

which equals to $\nabla_\theta J(\pi_\theta).$ A $Beta(\alpha, \beta)$ distribution on $[0, 1]$ has a well-known finite variance:
\[
\text{Var}(Z) = \frac{\alpha \beta}{(\alpha + \beta)^2 (\alpha + \beta + 1)}.
\]

After rescaling the beta range for action $[-h,h]$, 

\[
\text{Var}(A) = 4h^2 \, \text{Var}(Z) \leq 4h^2 \times \max_{Z \in \text{Beta}} \text{Var}(Z).
\]

And we already know $\text{Var}(Z) \leq \frac{1}{12}$ (if $\alpha = \beta = 1$, uniform). So:
\[
\text{Var}(A) \leq \frac{h^2}{3}.
\]

The variance of a rescaled Beta cannot exceed $h^3/3$, for all $\alpha,\beta>1$ (we assume the control policy should be unimodal).

\subsection{Realization Detail}

We assume control policy for locomotion scenario is unimodal, therefore, $\alpha,\beta>1$. We define the activation function for output layer of actor as shown below:

\[
SoftplusWithOffset(x)=Softplus(x)+1+1e-6
\]

During training, the action is sample from beta distribution and scale to $[-h,h]$. For deployment, we directly use the mean of distribution $\alpha/(\alpha+\beta)$ as the output.

\section{Network Architecture and Training Hyper-parameter}

In Table. \ref{tab:hyper}, we outline the hyperparameters for DHAL. Notably, the DHA is decoupled from both the encoder and the actor when ppo loss propagation and is only updated using the dynamics loss. During the PPO update, the PPO loss backpropagates through actor and encoder.

\begin{table}[]
\centering
\caption{Network Architecture and Training Hyper-parameter}
\begin{tabular}{ll}
\toprule
\textbf{Network Hyperparameters} & \textbf{value}\\
\midrule
\textit{DHA Architecture} & MLP\\
\textit{DHA Hidden Dims} & [256, 64, 32]\\
\textit{VAE Encoder Architecture} & 1-D CNN\\
\textit{VAE Encoder time steps} & 20\\
\textit{VAE Encoder Convolutional Layers} & Input channel = [30, 20] \\
\textit{VAE Encoder Convolutional Layers} & Kernel=(6,4), Stride=(2,2)\\
\textit{VAE Decoder Hidden Dims}  & [256, 128, 64]\\
\textit{VAE Latent Dims}  & 20\\
\textit{VAE KL Divergence Weight($\beta$)}  & 1e-2\\
\textit{Actor Hidden Dims}  & [512, 256, 128]\\
\textit{Gliding Critic Hidden Dims}  & [512, 256, 128]\\
\textit{Pushing Critic Hidden Dims}  & [512, 256, 128]\\
\textit{Sim2Real Critic Hidden Dims}  & [512, 256, 128]\\
\midrule

\textbf{PPO HyperParameters} & \textbf{Weight}\\
\midrule
\textit{Environments}  & 4096\\
\textit{Collection Steps} & 24\\
\textit{Discount Factor}  & 0.99\\
\textit{GAE Parameter}  & 0.9\\
\textit{Target KL Divergence}  & 0.01\\
\textit{Learning Rate Schedule}  & adaptive\\
\textit{Number of Mini-batches}  & 4\\
\textit{Clipping Paramete}  & 0.2\\

\bottomrule
\end{tabular}
\label{tab:hyper}
\end{table}

\section{Environment Setting and Sim2Real Detail}

\begin{table}[t]
\centering
\caption{Randomization and Noise}
\begin{tabular}{ll}
\toprule
\textbf{Property Randomization} & \textbf{value}\\
\midrule
\textit{Friction} & [0.6, 2.]\\
\textit{Added Mass} & [0, 3]kg\\
\textit{Added COM} & [-0.2, 0.2]\\
\textit{Push robot} & 0.5m/s per 8s\\
\textit{Delay} & [0, 20]ms \\
\midrule

\textbf{Sensor Noise} & \textbf{Weight}\\
\midrule
\textit{Euler Angle}  & $\mathcal{N} * 0.08$\\
\textit{Angular Velocity} & $\mathcal{N} * 0.4$\\
\textit{Projected Gravity}  & $\mathcal{N} * 0.05$\\
\textit{Joint Position}  & $\mathcal{N} * 0.05$\\
\textit{Joint Velocity}  & $\mathcal{N} * 0.1$\\

\bottomrule
\end{tabular}

\label{tab:hyper}
\end{table}

\subsection{Rolling Friction}

We observed that in Isaac Gym, the simulation of rolling objects, particularly wheels, is not highly accurate. This is primarily reflected in the discrepancies between simulated and real-world rolling friction, as well as imprecise collision detection. To address this, we applied a compensation force during training to roughly approximate the effects of rolling friction on different terrains for the forward and backward motion of the skateboard. The compensation force is defined by the following formula:

\[
F_{\text{push}, x} =
\begin{cases}
F_{\text{rand}}, & \text{if } v_{\text{skate}, x} > 0.3 \\
- F_{\text{rand}}, & \text{if } v_{\text{skate}, x} < -0.3 \\
0, & \text{otherwise}
\end{cases}
\]

\[
F_{\text{rand}} \sim U(10, 25)
\]

\[
v_{\text{skate}, x} = \text{quat\_rotate\_inverse} \left( q_{\text{skate}}, v_{\text{skate}} \right)
\]

Where\( F_{\text{push}, x} \) is the applied push force along the \( x \)-axis, \( F_{\text{rand}} \sim U(10, 25) \) is the randomly sampled force, and \( v_{\text{skate}, x} \) is the skateboard's velocity in its local frame, computed using the inverse quaternion rotation \( \text{quat\_rotate\_inverse}(q_{\text{skate}}, v_{\text{skate}}) \).

\begin{table}
\centering
\caption{Reward weights for two single-critic method(w-transfer/ wo-transfer )}
\begin{tabular}{lll}
\toprule
\textbf{Gliding Critic Reward} & \textbf{Weight} & \textbf{Weight(Transfer)}\\
\midrule
\textit{Feet on board} & 0.3 & $0.35 * 0.3$\\
\textit{Contact number} & 0.3 & $0.35 * 0.3$\\
\textit{Feet distance}  & 1.8 & $0.35 * 1.8$ \\
\textit{Joint positions}  & 1.2 & $0.35 * 1.2$\\
\textit{Hip positions}  & 1.2 & $0.35 * 1.2$\\
\midrule

\textbf{Pushing Critic Reward} & \textbf{Weight} & \textbf{Weight(Transfer)}\\
\midrule
\textit{Tracking linear velocity}  & 1.6  & $0.4 * 1.6$\\
\textit{Tracking angular velocity}  & 0.8  & $0.4 * 0.8$\\
\textit{Hip positions}  & 0.6  & $0.4 * 0.6$\\
\textit{Orientation}  & -2  & $0.4 * -2$\\
\midrule

\textbf{Sim2Real Critic Reward} & \textbf{Weight} & \textbf{Weight(Transfer)}\\
\midrule
\textit{Wheel contact number}  & 0.8  & $0.25 * 0.8$\\
\textit{Board--body height}  & 1  & $0.25 * 1$\\
\textit{Joint acceleration}  & -2.5e-7  & $0.25 * -2.5e-7$\\
\textit{Collisions}  & -1  & $0.25 * -1$\\
\textit{Action rate}  & -0.22  & $0.25 * -0.22$\\
\textit{Delta torques}  & -1.0e-7  & $0.25 * -1.0e-7$\\
\textit{Torques}  & -1.0e-5  & $0.25 * -1.0e-5$\\
\textit{Linear velocity (z-axis)}  & -0.1  & $0.25 * -0.1$\\
\textit{Angular velocity (x/y)}  & -0.01  & $0.25 * -0.01$\\
\textit{Base orientation}  & -25  & $0.25 * -25$\\

\bottomrule
\end{tabular}

\label{tab:singlecritic}
\end{table}

\subsection{Skateboard Truck model}

To simulate a realistic skateboard, we incorporated a bridge structure into the robotic skateboard, consisting of a front bridge and a rear bridge. Each bridge is modeled using a position-based PD controller to emulate spring dynamics, with the desired position and velocity set to zero at all times.

\subsection{Contact Detection}

We found that in Isaac Gym, the collision calculations for skateboard motion are imprecise. This is evident in the inaccuracies of the collision forces for passive rolling wheels as well as the collision forces between the robot and the skateboard. To address this, we designed the reward function to combine relative position error with collision forces to determine whether contact has occurred. This logic requires manual design and implementation.

\begin{figure}[h]
    \centering
    \includegraphics[width=1\linewidth]{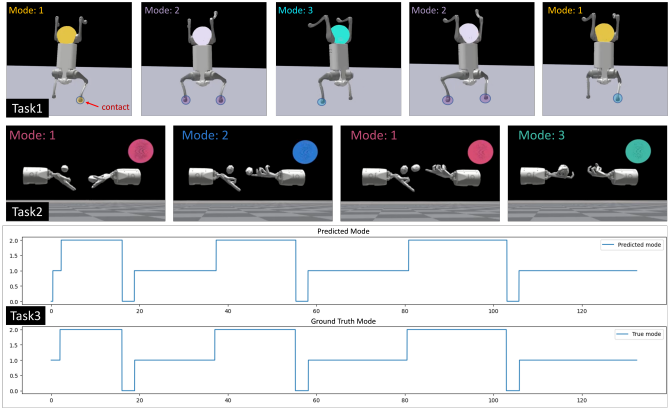}
    \caption{Validation on other hybrid tasks and mode Identification.}
    \label{fig:generalization}
\end{figure}

\section{Experimental Supplementary Notes}

\subsection{Training Cost}

We train policy on an NVIDIA RTX 3090, each iteration takes $3 \sim 4$ sec. Traning a policy could deploy in real-world costs $6 \sim 7$ hours in total.  

\subsection{Terrain setting}
We trained our policy directly on flat ground, yet it demonstrated the ability to generalize to challenging terrains such as skateboard parks in real-world scenarios. At first, we also tried training it on complex terrain by including stairs and slopes. We observed no obvious advantage, while the training time increased. Therefore, the policies are all trained on flat ground.

\subsection{Single-Critic Reward}

In the experimental section, we compared the performance of single-critic and multi-critic approaches. To help readers clearly understand the differences between the two single-critic setups, we list their respective reward configurations in Table. \ref{tab:singlecritic}.

\subsection{Extreme Terrain Experiments}
\label{app:exp}
As shown in Fig. \ref{fig:skbp}, we tested our robot in a skatepark specifically designed for extreme skateboarding, featuring challenging multi-level stair sets, U-shaped bowls, and terrain with cliff-like characteristics. This environment allowed us to evaluate the limits of our algorithm’s performance. Surprisingly, our algorithm remained relatively stable across these complex terrains. Although the robot occasionally deviated from the skateboard due to terrain disturbances, it was still able to recover and maintain a graceful skating posture.

\subsection{Hybrid Dynamics automata Validation in other task}

We apply the proposed framework to more tasks to demonstrate its effectiveness. \textbf{1) Robot Hand-stand Locomotion, 2) Dexterous Manipulation Hand-over, 3) Switching Linear Dynamical System} shown in Fig. \ref{fig:generalization}. (1-2) aim at contact-rich problems, and 3) is a classic textbook example in hybrid dynamical systems with verifiable ground truth. In 1), we can identify the gait or contact mode. In 2), we can identify different stages of the object hand-over task: in the air, pushed by the right hand and caught by the left hand. In 3), we can compare our identification with the analytical solution.

\end{appendices}

\end{document}